\documentclass[letterpaper, 10 pt, conference]{ieeeconf}  

\usepackage{xcolor}
\usepackage{amsmath}
\usepackage{amsthm}
\usepackage{amsfonts,amssymb,mathtools}

\usepackage{cleveref}

\newcommand{\HH}{\mathcal{H}}

\newcommand{\KK}{\mathcal{K}}

\newcommand{\NN}{\mathcal{N}}

\newcommand{\XX}{\mathcal{X}}

\newcommand{\real}{\mathbb{R}}

\newcommand{\st}{\textit{s.t. }}

\DeclareMathOperator*{\argmax}{arg\,max}

\crefname{asp}{assumption}{assumptions}
\Crefname{asp}{Assumption}{Assumptions}
\crefname{lem}{lemma}{lemmas}
\Crefname{lem}{Lemma}{Lemmas}

\theoremstyle{definition}
\newtheorem{thm}{Theorem}
\newtheorem{lem}[thm]{Lemma}

\newtheorem*{cor}{Corollary}

\DeclareMathOperator{\interior}{int}

\newtheorem{asp}{Assumption}
\newtheorem{algo}{Algorithm}

\theoremstyle{definition}
\newtheorem{defn}{Definition}

\theoremstyle{remark}
\newtheorem*{rem}{Remark}

\theoremstyle{remark}

\newcommand{\re}[1]{{\color{black} #1}}
\newcommand{\rre}[1]{{\color{black} #1}}

\usepackage[utf8]{inputenc}
\usepackage{comment}
\usepackage{subfigure}
\usepackage{graphicx}
\usepackage[normalem]{ulem}
\usepackage{booktabs}
\usepackage{floatrow}

\usepackage{cleveref}

\IEEEoverridecommandlockouts                              

\title{\LARGE \bf
\re{Persistently Feasible Robust Safe Control \\by Safety Index Synthesis and Convex Semi-Infinite Programming
}}

\author{Tianhao Wei$^{1}$, Shucheng Kang$^{2}$, Weiye Zhao$^{1}$, and Changliu Liu$^{1}$
\thanks{*This material is based upon work supported by the National Science Foundation under Grant No. 2144489.}
\thanks{$^{1}$These authors are with Robotics Institute, Carnegie Mellon University, {\tt\small twei2, weiyezha, cliu6@andrew.cmu.edu}}%
\thanks{$^{2}$Shucheng Kang is with the Department of Electrical Engineering, Tsinghua University. Work done during an internship at Carnegie Mellon.
        {\tt\small ksc19@mails.tsinghua.edu.cn}}%
}

\begin{document}

\maketitle
\thispagestyle{empty}
\pagestyle{empty}


\begin{abstract}
Model mismatches prevail in real-world applications. Ensuring safety for systems with uncertain dynamic models is critical. \re{However, existing robust safe controllers may not be realizable when control limits exist. And existing methods use loose over-approximation of uncertainties, leading to conservative safe controls. To address these challenges, we propose a control-limits aware robust safe control framework for bounded state-dependent uncertainties. 
We propose safety index synthesis to find a robust safe controller guaranteed to be realizable under control limits.
And we solve for robust safe control via Convex Semi-Infinite Programming, which is the tightest formulation for convex bounded uncertainties and leads to the least conservative control. 
In addition, we analyze when and how safety can be preserved under unmodeled uncertainties. Experiment results show that our robust safe controller is always realizable under control limits and is much less conservative than strong baselines.
}
\end{abstract}

\section{Introduction}

Safety is critical in robotic systems. Safe control, as the last defense of the system, ensures real-time safety by keeping the system state in a safe set (known as forward invariance)~\cite{liu2014control}. 
Energy-function based methods were proposed to realize safe control~\cite{wei2019safe}.
With the energy function (also called safety index, barrier function, and Lyapunov functions), the safe control problem is converted to online quadratic programming (QP). 
QP-based safe control has been widely studied for deterministic dynamic models~\cite{wei2019safe}.

However, dynamic models are approximations of the real-world systems~\cite{wajid2022Formal}. There can always be uncertainties. \re{Safe control has been extended to \textit{uncertain dynamic models} (UDM)~\cite{taylor2021towards, castaneda2021pointwise}}. 
\re{However, existing works usually assume the robust safe control is always realizable (also called \textit{persistently feasible})~\cite{buch2021robust, garg2021robust} or there is no control limit~\cite{jankovic2018robust, taylor2021towards, castaneda2021pointwise}. 
It remains unclear how to design a persistently feasible robust safe controller under control limits.}

Another challenge of robust safe control is how to solve the QP-based safe control with model uncertainty and control limits. There can be infinitely many constraints in such a QP because each possible dynamic model leads to one constraint. \re{Therefore, existing methods use different ways to reduce the constraints by over-approximating the uncertainties.} 
Some methods upper bound the residual time derivative of the barrier function induced by uncertainty~\cite{cosner2021MeasurementRobust, nguyen2022Robust, brunke2022Barrier}. However, sometimes it is unclear how to choose the bound. And the bound is a coarse over-approximation because it is often state-independent or control-independent, leading to conservative safe controls. 
\re{Some methods formulate robust safe control as Second Order Cone Programming (SOCP)~\cite{taylor2021towards, castaneda2021pointwise, grover2022control}. However, as we will show in the experiment (\cref{fig: SCARA-CR-VS-GR} and \cref{fig: Uncertainty_bounds_and_safe_control_set}), SOCP formulation can greatly over-approximate the uncertainty leading to infeasible control, or lead to safety violations when the uncertainty is not Gaussian.} 
\re{\rre{The presence of control limits requires a tight approximation of the uncertainty}. Otherwise, the desired control may exceed the control limits due to over-approximation of the uncertainty.}
Besides, the modeled uncertainty may not be accurate. It is unclear what happens when the actual uncertainty is greater than we expected.

To address the above challenges, we propose a general robust safe control framework. First, we propose a sound but not complete method \textit{UR-SIS} (uncertainty-robust safety index synthesis) to synthesize a control-limit aware and uncertainty-robust safe controller that is guaranteed to be persistently feasible. In case such a safety index does not exist or we cannot find one, UR-SIS tries to maximize feasible state rate.
\re{Second, we formulate robust safe control with model uncertainty as  \textit{convex semi-infinite programming} (CSIP), which is the \textbf{exact} formulation (no over-approximation) for convex bounded uncertainties and the \textbf{tightest} convex formulation for arbitrary bounded uncertainties.} We propose a real-time optimal solver based on the cutting-plane method. We also show that, \re{SOCP formulation is a special case of our CSIP formulation for hyper-ellipsoid bounded uncertainties.} Compared to existing methods, our method considers exact or tight uncertainty bounds, therefore \re{leads to larger admissible control sets and consequently less-conservative robust safe controls}.
Lastly, we provide a theoretical analysis of what happens when unmodeled uncertainty exists. In particular, we present the condition on the unmodeled uncertainty that the system state can be kept in a larger set (than the safe set) with the learned UR-SI. Combining all the parts, we are able to ensure robustness and non-conservativeness of the safe control, and forward invariance of the safe set for bounded dynamic model uncertainties and even unmodeled uncertainties.


The remainder of the paper is organized as follows: 
\Cref{sec: formulation} formulates the problem and introduces notations. 
\Cref{sec: method} presents the proposed robust safe control framework in detail. 
\Cref{sec: exp} evaluates the proposed methods on two examples: \re{planar robot} and Segway.

\section{Formulation}\label{sec: formulation}

We consider the following nonlinear control-affine system with uncertain dynamic models (UDM)~\footnote{Although our method assumes control affine dynamics, it is applicable to non-control affine systems, since we can always have a control affine form through dynamics extension~\cite{liu2016algorithmic}.}:
\begin{align}
    \dot x = f(x) + g(x) u,\ f(x) \in \Sigma_f(x),\ g(x) \in \Sigma_g(x), \label{eq: dyn}
\end{align}
where state $x \in X \subseteq \real^{n}$ and control $u \in U \subseteq \real^{m}$. 
The terms $f(x) \in \real^{n}$ and $g(x) \in \real^{n\times m}$ are random matrices (elements $f(x)_{i}, g(x)_{i,j}$ are random variables) which 
are bounded by $\Sigma_f(x) \subseteq \real^{n}$ and $\Sigma_g(x) \subseteq \real^{n\times m}$ respectively. $\Sigma_f(x)$ and $\Sigma_g(x)$ are measurable functions on the state space which are state-dependent. 
For simplicity, in the following discussion, we may write $\forall f(x), g(x)$  to represent the condition $\forall f(x) \in \Sigma_f(x),\ g(x) \in \Sigma_g(x)$. 


We consider the safety specification as a requirement that the system state should be constrained in a closed and connected set $\XX_S \subseteq X$, which we call the safe set. We assume $\XX_S$ is the zero-sublevel set of a \textit{safety index} $\phi_0: X \mapsto \real$ given by the user. That is $\XX_S = \{x \mid \phi_0(x) \leq 0, x \in X\}$.
Constraining states inside $\XX_S$ can be expressed as a \textit{forward invariance} problem: when $\phi_0(x(t_0)) \leq 0$, ensure $\phi_0(x(t)) \leq 0,\ \forall t>t_0$. Forward invariance can be guaranteed with minimal invasion by QP-based safe set algorithms.
\begin{algo}[QP-based safe set algorithms]\label{algo: ssa}
Given a reference control $u_{\text{ref}}$ and a safety index $\phi$, QP-based safe set algorithms find safe control $u$ by: 
\begin{align}
    \min_u \|u - u_{\text{ref}}\|^2\
    \st \dot \phi(x,u) \leq -\gamma(\phi(x)),\label{eq: ssa}
\end{align}
where $\gamma$ is a piecewise smooth function and $\gamma(\rre{\phi(x)}) > 0$ when $\phi(x) > 0$. $\gamma$ can be non-continuous and designed differently.
A special case of $\gamma$ is an extended class $\KK$ function on $\phi(x)$, corresponding to the control barrier function (CBF) method~\cite{wei2019safe}. 
Safe set algorithms require $\phi$ to be \textit{persistent feasible}: $\forall x, \exists u$, such that $\dot \phi(x,u) \leq -\gamma(\rre{\phi(x)})$. However, a user-defined safety index $\phi_0$ may not be naturally persistently feasible. It is often necessary to design a $\phi$ based on $\phi_0$ such that we can constrain the state in $S: \{\phi \leq 0\} \subseteq \XX_S$~\cite{liu2014control}. 
\end{algo}




Extending \cref{algo: ssa} to UDM requires the safety constraint holds for all possible models. We define the extended problem as \textit{uncertainty-robust QP} (UR-QP).
\begin{defn}[UR-QP]
\begin{align}
    \min_u  \|u - u_{\text{ref}}\|^2\  \st \dot\phi(x,u) \leq -\gamma(\phi(x)),\  \forall f(x), g(x). \label{eq: rssa}
\end{align}
\end{defn}


And we say $\phi$ is an \textit{uncertainty-robust safety index} (UR-SI) if it ensures persistent feasibility for all possible models.

\begin{defn}[UR-SI]
A safety index $\phi$ is a UR-SI if there exists a piecewise smooth, strictly increasing function $\gamma$, and $\gamma(0)=0$, such that  the following \textit{feasibility} condition holds: 
\begin{align}
    \forall x, \exists u,\ \dot \phi(x,u) \leq -\gamma(\phi(x)),\ \forall f(x), g(x) \label{eq: robust_feasibility}
\end{align}
\end{defn}

To solve the UR-QP \cref{eq: rssa} or to verify \cref{eq: robust_feasibility}, we are interested in the robust safe control set $U_r(x):$
\begin{align}
    U_r(x) : = \{u \mid \dot\phi(x,u) \leq -\gamma(\phi(x)),\  \forall f(x), g(x)\}.\label{eq: uc_defn}
\end{align}
With $U_r(x)$, \cref{eq: rssa} can be converted into the following form
\begin{align}
    & \min_{u\in U \cap U_r(x)} \|u - u_{\text{ref}}\|^2. \label{eq: uc_rssa}
\end{align}

Prior work~\cite{noren2021safe} has shown that $U_r(x)$ can be defined with the Lie derivatives. For simplicity, we may omit $(x)$ when there is no ambiguity.

\begin{defn}
We denote Lie derivatives by $L_f\phi \coloneqq \nabla \phi^T f$, $L_g\phi \coloneqq \nabla \phi^T g$.
And the range of $L_f\phi$ and $L_g\phi$ are denoted by $V_f \coloneqq \{L_f\phi \mid f \in \Sigma_f\}$ and $\ V_g \coloneqq \{L_g\phi \mid g \in \Sigma_g\}$.
\end{defn}
Then the safety constraint can be transformed as follows:
\begin{align}
     & \forall f, g, \dot\phi(x,u) \leq -\gamma(\phi) \iff \ \max_{f,g} \dot \phi(x,u) \leq -\gamma(\phi)\\
     & \iff \ \max_{L_f\phi, L_g\phi} L_f\phi + L_g\phi \cdot u \leq -\gamma(\phi) \label{eq: fg_joint}\\
     & \iff \max_{L_g\phi \in V_g} L_g \phi \ u \leq -\gamma(\phi) - \max_{L_f\phi \in V_f} L_f\phi =: c \label{eq: rssa_max}
\end{align}
Then $U_r\re{(x)}$ can be defined equivalently by
\begin{align}
    U_r = \{u \mid \max_{v \in V_g} v^T u \leq c\} = \{u \mid v^T u \leq c, \forall v \in V_g\}. \label{eq: uc_convex}
\end{align}
Although $U_r$ is defined by linear constraints, the lack of a closed form of $U_r$ makes it difficult to solve \cref{eq: uc_rssa}. Even deciding if $U_r$ is empty is difficult. Prior work~\cite{noren2021safe} considers a simplified scenario and proposes a minimax formulation. But the minimax is generally intractable. In this work, we study how to solve this problem precisely and efficiently.

Besides, it is challenging to synthesize a UR-SI. Because a UR-SI requires \cref{eq: robust_feasibility} to hold for all $x$, which involves infinitely many states. And, how to deal with unmodeled uncertainty remains unknown.

\section{Method}\label{sec: method}

To address these challenges, we first introduce a method to synthesize a UR-SI with finite states. Then we present \re{an efficient and tight solver of the UR-QP by CSIP}, which give optimal solutions of \cref{eq: uc_rssa} for convex bounded uncertainties. \re{We also show that when the bound is an ellipsoid, CSIP reduces to SOCP.}
 In the end, we analyze how the forward invariance set changes in presence of unmodeled large uncertainty.

\subsection{Uncertainty-robust safety index synthesis (UR-SIS)}
The purpose of safety index synthesis is to design a UR-SI $\phi$ based on $\phi_0$ such that the closed-loop trajectory is constrained in the safe set $\XX_S$ even under uncertainty. 

To synthesize a UR-SI, we extend our previous work on synthesizing safety index for deterministic learned dynamics~\cite{wei2022safe}. We proved that the feasibility of the whole state space can be ensured by only verifying finite sampled states. In this work, \re{we generalize the method to uncertain dynamics and adjust the sampling density based on the uncertainty. 
Then we will be able to use an evolutionary algorithm to find a valid UR-SI with the sampled states. This method works for low-dimensional parameterizations of CBF, which we call the safety index~\cite{liu2014control}.}



We first prove that $\phi$ is a UR-SI if it has feasible solutions on a sampled state set $B$ based on the following assumption: 
\begin{asp} \label{asp: lipschitz}
$\gamma$, $\phi$ and $\nabla \phi$ are Lipschitz continuous functions with Lipschitz constants  $k_\gamma$, $k_\phi$ and $k_{\nabla \phi}$ respectively. The set change $\Delta(\Sigma_f, x, x')$ and $\Delta(\Sigma_g, x, x')$ are bounded by $k_{\Sigma_f} \|x-x'\|$ and $k_{\Sigma_g} \|x-x'\|$ respectively, where $\Delta(A, x, x') : = \max_{a\in A(x)} \min_{a'\in A(x')} \|a-a'\|$.
And $\|u\|$ and $\|\dot x\|$ are bounded by $M_u$ and $M_{\dot x}$ respectively.
\end{asp}
\begin{lem}
Suppose 1) we sample a state subset $B \subset X$ such that  $\forall x \in X$, $\min_{x' \in \re{B}} \|x - x'\| \leq \delta$, where $\delta$ is a constant representing the sampling density; 2) $\forall x' \in B$, there exists a safe control $u$, \st $\dot \phi(x', u) \leq - \gamma(\phi(x')) -\epsilon,\  \forall f(x') , g(x')$,
where $\epsilon = k_{\phi} (k_{\Sigma_f} + k_{\Sigma_g} M_u)\delta  + k_{\nabla \phi} \delta M_{\dot x} + k_\gamma k_\phi \delta$. Then $\phi$ satisfies \cref{eq: robust_feasibility} and therefore is a UR-SI.

\begin{proof}
According to condition 1), $\forall x\in X$, $\exists x'\in B$ such that $\|x-x'\|\leq \delta$.  And based on \cref{asp: lipschitz}, we have
$\|f(x) - f(x')\| \leq k_{\Sigma_f} \delta$, and $\|g(x) - g(x')\| \leq k_{\Sigma_g} \delta$. According to condition 2), for this $x'$, we can find $u$ s.t. $\dot \phi(x', u) \leq - \gamma(\phi(x')) -\epsilon$. Next we show $x$ and $u$ satisfy \cref{eq: robust_feasibility} by \underline{triangle inequality} and \dashuline{Lipschitz condition}:
\re{
\begin{align}
    & \dot \phi(x,u) = \dot \phi(x,u) - \dot \phi(x',u) + \dot \phi(x',u)\\
= & \nabla \phi(x) \dot x - \nabla \phi(x') \dot x' + \dot \phi(x',u)\\
= & \nabla \phi(x) \dot x  - \nabla \phi(x') \dot x'  - \nabla \phi(x) \dot x'  + \nabla \phi(x) \dot x'  + \dot \phi(x',u) \\
= & \nabla \phi(x) (\dot x - \dot x')  + [\nabla \phi(x) - \nabla \phi(x')] \dot x'  \nonumber \\ 
& -\gamma(\phi(x)) + \gamma(\phi(x)) + \dot \phi(x',u) \label{eq:before_tri} \\
\leq & \underline{\|\nabla \phi(x)\| \|f(x) + g(x) u - f(x') - g(x')u\|} \nonumber \\
& + \underline{\|\nabla \phi(x) - \nabla \phi(x')\| \|\dot x'\|} \nonumber \\ 
& -\gamma(\phi(x)) + \gamma(\phi(x)) - \gamma(\phi(x'))-\epsilon \label{eq:before_lip} \\
\leq & \dashuline{k_\phi (k_{\Sigma_f} \delta + k_{\Sigma_g} \delta M_u)} + \dashuline{k_{\nabla \phi} \delta M_{\dot x}} \nonumber \\
    &  -\gamma(\phi(x)) \dashuline{+ k_\gamma k_\phi \delta} -\epsilon\\
\leq & - \gamma(\phi(x))
\end{align}
}
Therefore, $\phi$ ensures feasibility for all states.
\end{proof}
\end{lem}

With this lemma, we can validate if a given $\phi$ is a UR-SI with a finite sampled state set $B$. Next, we show how to find a UR-SI $\phi$ with $B$. Previous work \cite{liu2014control} has proposed a general parameterized form $\phi_{\theta}$ that guarantees $S: =\{\phi_{\theta}\leq 0\} \subseteq \XX_S$ without considering the feasibility of $\phi_\theta$. The parameterization also applies to UDM because $S$ only depends on the state, which is certain. Therefore, all we need to do is to optimize $\theta$ to make  $\phi_{\theta}$ a UR-SI. The synthesis problem can be formulated as 
\begin{align}
    \max_{\theta} \left|\{ x \mid \min_{u} \dot \phi_{\theta}(x,u) < -\gamma(\phi_{\theta}(x))\}\right|.
\end{align}
The problem is non-differentiable. Therefore we apply a derivative-free evolutionary algorithm CMA-ES\rre{~\cite{wei2022safe}}, which iteratively optimizes $\theta$ by evaluating current $\theta$ candidates on $B$ and proposing new $\theta$ candidates from the best performers. The objective is to maximize the feasible rate $r\coloneqq \frac{|B^*|}{|B|}$, where $B^* : = \{x' \mid \min_u \dot \phi(x', u) \leq - \gamma(\phi(x')) -\epsilon, \forall f(x') , g(x').\ x'\in B\}$ is the feasible sampled state set. 
To decide if a state $x'$ belongs to $B^*$, we rely on the robust safe control solvers to be proposed next.
The algorithm stops when the feasible rate $r$ converges. If $r=1$, then we find a UR-SI $\phi$. This method is sound but not complete. That is, we cannot assert a UR-SI does not exist if we cannot find one by this method. We will investigate complete methods in the future.

\subsection{Robust safe control for convex bounded uncertainties}\label{sec: convex_rssa}

The UR-QP \cref{eq: uc_rssa} is difficult to solve because the lack of the closed form of $U_r\re{(x)}$ makes it difficult to characterize the boundary of $U_r\re{(x)}$. To address this issue, our key observation is that when $\Sigma_g\re{(x)}$ is a polytope, $U_r\re{(x)}$ can be well over-approximated by a polytope set $\hat U_r\re{(x)}$ which has a closed form of boundary. And the over-approximation can be iteratively reduced to zero. If $\Sigma_g\re{(x)}$ is not a polytope, we can construct its convex hull and then apply the method. 

Next, we first introduce the polytope assumption and then give a formal description of the method. \re{Note that the method do not rely on \cref{asp: lipschitz}}.
\begin{defn}[$g_{\text{flat}}\re{(x)}$]
Represent $g\re{(x)}$ by $\left[ \overrightarrow{g_1},\overrightarrow{g_2},\cdots \overrightarrow{g_n} \right] \in  \real^{n\times m}$ ,where $\overrightarrow{g_i}\in \mathbb{R} ^m$. 
Then we define $g_{\text{flat}}\re{(x)}$ as $\left[ \overrightarrow{g_1}^T, \overrightarrow{g_2}^T, \cdots, \overrightarrow{g_n}^T \right]^T \in \mathbb{R} ^{nm}$. 
\end{defn}

\begin{asp}[Polytope $\Sigma_f\re{(x)}, \Sigma_g\re{(x)}$] \label{asp: convex_g}
We assume the uncertainties are bounded by polytopes $\Sigma_f\re{(x)}$ and $\Sigma_g\re{(x)}$. That is, $\exists A_f, b_f, A_g, b_g$ such that
    \begin{align}
        &\Sigma_f\re{(x)} = \{f\re{(x)} \mid A_f\re{(x)} f\re{(x)} < b_f\re{(x)}\},\\ 
        &\Sigma_g\re{(x)} = \{g\re{(x)} \mid A_g\re{(x)} g_{\text{flat}}\re{(x)} < b_g\re{(x)}\}.
    \end{align}
    \re{Consequently}, $V_f\re{(x)}$ and $V_g\re{(x)}$ are also polytopes because $\nabla \phi\re{(x)}$ is a linear transformation, 
\end{asp}


A polytope $\Sigma_f\re{(x)}$ eases the computation of $c\re{(x)}$ in the RHS of \cref{eq: rssa_max}. In this case, $c\re{(x)}$ can be solved by linear optimization because $V_f\re{(x)}$ is a linear convex set, as done in~\cite{liu2015safe}. 
Next, \rre{we propose a necessary condition of the existence of $U_r(x)$} and an efficient method to solve \cref{eq: uc_rssa}. 
\begin{lem}
    \rre{If $U_r(x)$ is not an empty set, then when $c(x) < 0$, $0 \notin \interior V_g(x)$.}

\begin{proof}
    We prove this by contradiction. $0 \in \interior V_g\re{(x)} \implies \exists r>0$, s.t. $v \in V_g\re{(x)}$ if $\|v\|<r$. Given a $v$ that $\|v\| < r$, suppose $U_r\re{(x)} \neq \emptyset$, then $\exists u\in U_r\re{(x)}$ such that $v^T u < c < 0$. But $-v \in V_g\re{(x)}$ (because $\|-v\|<r$) and $-v^T u > 0$, therefore $u\notin U_r\re{(x)}$, Contradicts with the assumption.
\end{proof}
\end{lem}

\begin{figure}
\vspace{7pt}
\floatbox[{\capbeside\thisfloatsetup{capbesideposition={right,top},capbesidewidth=0.49\linewidth}}]{figure}[\FBwidth]
{\caption{\small $U_r$ is defined by infinite constraints but can be over-approximated by the orange ones, which correspond to vertices of $V_g$. The purple constraint represents the one with the maximum violation and will be added to $\hat U_r$.}\label{fig: U_r}}
{\includegraphics[width=0.98\linewidth]{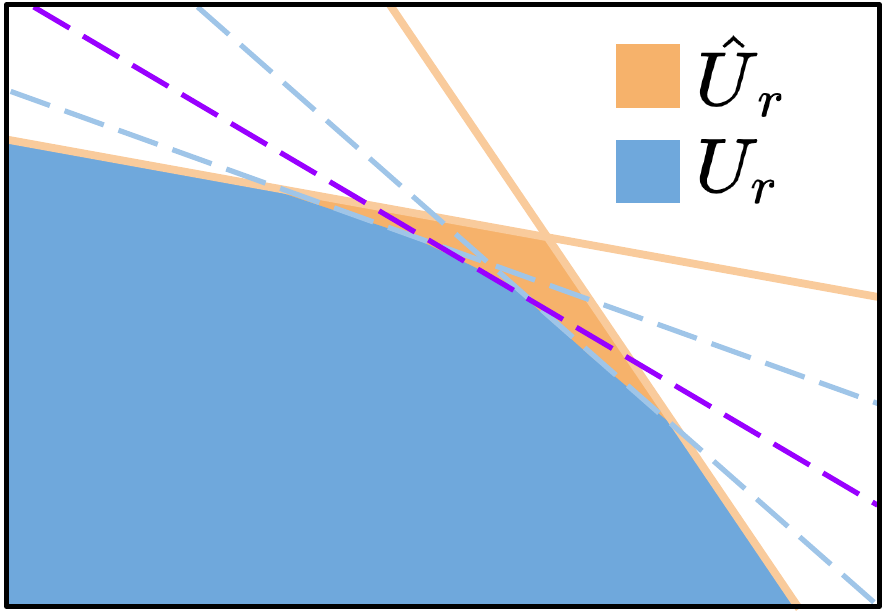}}
\vspace{-15pt}
\end{figure}

\begin{algo}[Polytope RSSA]\label{algo: convex_rssa}
Our key insight is that the boundary of $U_r\re{(x)}$ is mostly defined by $\HH$: the vertices of $V_g\re{(x)}$. We define  $\hat U_r\re{(x)} : = \{u \mid v^T u \leq c,  \forall v \in \HH \subseteq V_g\re{(x)}\}.$
It is easy to see that $U_r\re{(x)} \subseteq \hat U_r\re{(x)}$ as shown in \cref{fig: U_r}. To solve the UR-QP \cref{eq: uc_rssa}, we first minimize $\|u - u_{\text{ref}}\|$ in $U\re{(x)} \cap \hat U_r\re{(x)}$, which is a QP with finite linear constraints. \re{Then, we find the constraint of max violation: $v_* = \argmax_{v \in V_g(x)} v^T u$. If $v_*^Tu > c $, we add $v_*^Tu \leq c$ to $\hat U_r(x)$ and repeat the process}.
\end{algo}
\begin{lem}
\Cref{algo: convex_rssa} converges to the optimal solution.

\begin{proof}
\Cref{eq: uc_rssa} is a special case of convex semi-infinite programming (CSIP), and Polytope RSSA is a variant of the cutting-plane method~\cite{gustafson1973numerical}. It has been proved that the method always converges to an optimal solution in~\cite{gustafson1973numerical}.
\end{proof}
\end{lem}


\re{A special case of Polytope RSSA is when $\Sigma_f(x)$ and $\Sigma_g(x)$ are hyper-ellipsoids. In this case, CSIP can be reduced to Second Order Cone Programming (SOCP), which is particularly useful when we consider chance constraints for Gaussian uncertainties. SOCP can be solved efficiently by Semi-Definite Programming (SDP) solvers.}


\begin{asp}[Hyper-ellipsoid $\Sigma_f\re{(x)}$, $\Sigma_g\re{(x)}$] \label{asp: gau_g}
$\exists \mu_f\re{(x)}\in \real^{n}$, $Q_f\re{(x)}\in \real ^{n\times n},Q_f\succ 0$, $d_f\re{(x)} \in \real^+$, and $\exists \mu_g\re{(x)}\in \real^{mn}$, $Q_g\re{(x)}\in \real^{mn\times mn}$, $Q_g\succ 0$, $d_g\re{(x)} \in \real^+$, \st
\begin{align}
    \Sigma _f\re{(x)} & \coloneqq \{ f \mid (f-\mu_f )^T Q_{f}^{-1} ( f-\mu_f ) \leqslant d_f \},\\
    \Sigma _g\re{(x)} & \coloneqq \{ g_{\text{flat}} \mid ( g_{\text{flat}}-\mu _g ) ^TQ_{g}^{-1}( g_{\text{flat}}-\mu _g ) \leqslant  d_g\}.
\end{align}
\end{asp}

\begin{rem}
If $f \sim \NN \left( \mu_f,Q_f \right)$, $g_{\text{flat}}\sim \NN \left( \mu _g,Q_g \right)$, $d_f=\chi_n^{2}(\alpha)$ and $d_g=\chi_{mn}^{2}(\alpha)$, where $\alpha \in \real^+$ and $\chi^2$ is the chi-square function. Then the above hyper-ellipsoids correspond to the $1-\alpha$ confidence bounds.


\end{rem}


\begin{cor}
$L_g\phi$, $L_f\phi$ are also bounded by hyper-ellipsoids. Let $\mu_v\re{(x)} = \nabla \phi \cdot \mu_g$,  $Q_v\re{(x)} = \nabla \phi^T Q_g \nabla \phi$. Then
\begin{align}
    V_g\re{(x)} \coloneqq \{ v \mid ( v-\mu _v ) ^TQ_{v}^{-1}( v-\mu _v ) \leqslant d_g \}.
\end{align}
Similarly, we can define $V_f\re{(x)}$ and compute $c$ as in \cref{eq: rssa_max}.
\end{cor}

The lemma below gives us an equivalent form of $U_r\re{(x)} \coloneqq\{ u \mid v^Tu\leqslant c,\forall v\in V_g\re{(x)} \}$ using the hyper-ellipsoid bounds, which is defined by only one constraint. 

\begin{algo}[Ellipsoid RSSA]\label{algo: ellipsoid_rssa}
Under \cref{asp: gau_g}, $\re{\forall x},\exists L, \st L L^T = d_g Q_v$. Therefore $U_r\re{(x)}$ has the second-order cone form and makes the UR-QP \cref{eq: uc_rssa} a SOCP:
\begin{align}
   U_r\re{(x)}=\{ u \mid ||L^T u||\leqslant -\mu_v^T u + c \} 
\end{align}


\begin{proof}
For simplicity, let $Q\re{(x)}:=d_g\re{(x)} Q_v\re{(x)} $. Then, the hyper-ellipsoid can be described as 
\begin{align}
    \Sigma _v\re{(x)} = \{v \mid ( v-\mu_v) ^TQ^{-1} ( v-\mu_v ) \leqslant 1\}
\end{align}

Because $Q\succ 0$, there exists an invertible matrix $L$, s.t. $Q=LL^T$. Define $v'=L^{-1}( v-\mu_v)$, then $V_g$ could be transformed to $V_g'$: $V_g' = \{v' \mid v'^Tv'\leqslant 1\}$.
Substitute $v$ with $v'$ in $U_r$, we could get
$$
    U_r\re{(x)}=\{ u \mid ( Lv'+\mu_v) ^Tu\leqslant c,\forall v'\in V_g' \}
$$
To further simplify $U_r$'s form, we first introduce an auxiliary variable $u'$ s.t. $u=( L^T ) ^{-1}u'$. 
Notice that $\underset{v'^Tv'\leqslant 1}{\max}v'^Tu'=||u'||$.
Therefore, 
\begin{align}
    U_r\re{(x)}&=\{ ( L^T ) ^{-1}u' \mid ||u'||+( L^{-1}\mu_v) ^Tu'\leqslant c \}\\
    &=\{ u \mid ||L^T u||\leqslant -\mu_v^T u + c \},
\end{align}
which is a second-order cone form.
\end{proof}
\end{algo}


\subsection{Forward invariance under unmodeled uncertainty}

We have discussed how to design a robust safety index and how to find robust safe control with known uncertainty ranges. However, in practice, sometimes we cannot acquire an accurate uncertainty range in advance. Further analysis for unmodeled uncertainties is desired.
\rre{We present the condition of the unmodeled uncertainties such that the forward invariance is still preserved (in a larger set) by a UR-SI}.

\begin{defn}[Residual dynamics]
We define the unmodeled uncertainty as unknown residual dynamics $\Tilde{f}(x)$ and $\Tilde{g}(x)$: 
\begin{align}
\begin{split}
    \dot x =& f(x) + g(x) u + \Tilde{f}(x) + \Tilde{g}(x) u,\\
    &f(x) \in \Sigma_f(x), g(x) \in \Sigma_g(x) \label{eq: actual_dyn}
\end{split}
\end{align}
\end{defn}

    

\begin{lem} \label{thm: FI_set}
A UR-SI $\phi$ ensures the forward invariance of the set $\hat S = \{x \mid \phi(x) \leq \gamma^{-1}(k_{\phi} M_{\Tilde{\dot x}})\}$ with unmodeled uncertainties shown in \cref{eq: actual_dyn}, where $M_{\Tilde{\dot x}} : = \max_{x,u} \|\Tilde{f}(x) + \Tilde{g}(x) u\|$ represents the maximum norm of the residual dynamics: 

\begin{proof}
It suffices to show that applying $\phi$ on the nominal dynamics \cref{eq: dyn} ensures $\dot \phi \leq 0$ on $\partial \hat S := \{x \mid \phi(x) = \gamma^{-1}(k_{\phi} M_{\Tilde{\dot x}})\}$ for the true dynamics \cref{eq: actual_dyn}. 

$\dot \phi$ can be decomposed as $\dot \phi = \dot \phi_{\text{nom}} + \dot \phi_{\text{res}}$, where $\dot \phi_{\text{nom}} := \nabla \phi [f(x) + g(x) u]$, $\dot \phi_{\text{res}} := \nabla \phi(x) [\Tilde{f}(x) + \Tilde{g}(x) u]$.
Because $\forall x, \exists u, \dot \phi_{\text{nom}}(x) \leq -\gamma(\phi(x)), \forall f(x), \forall g(x)$ and
\begin{align}
    \dot \phi_{\text{res}}(x) & = \nabla \phi(x) [\Tilde{f}(x) + \Tilde{g}(x) u]\\
    &\leq k_{\phi} \|\Tilde{f}(x) + \Tilde{g}(x) u\| \leq k_{\phi} M_{\Tilde{\dot x}}.
\end{align}
We have $\forall x \in \partial \hat S, \exists u$ such that $\forall f(x), \forall g(x)$
\begin{align}
    \dot \phi(x) &= \dot \phi_{\text{nom}}(x) + \dot \phi_{\text{res}}(x) \leq -\gamma(\phi(x)) + k_{\phi} M_{\Tilde{\dot x}}\\
    &= -\gamma(\gamma^{-1}(k_{\phi} M_{\Tilde{\dot x}})) + k_{\phi} M_{\Tilde{\dot x}} = 0
\end{align}
Therefore, $\hat S$ is forward invariant with the UR-SI $\phi$.
\end{proof}
\end{lem}

\section{Experiment} \label{sec: exp}

Our method applies to arbitrary dynamic model uncertainties as long as they can be bounded by state-dependent polytopes or ellipsoids, including but not limited to Gaussian Process, neural network dynamic models, and parameterized models. For simplicity, we show the effectiveness of our method on two parameterized models with non-Gaussian and Gaussian uncertainties.


\begin{figure}
    \vspace{5pt}
    \centering
    \includegraphics[width=0.7\linewidth]{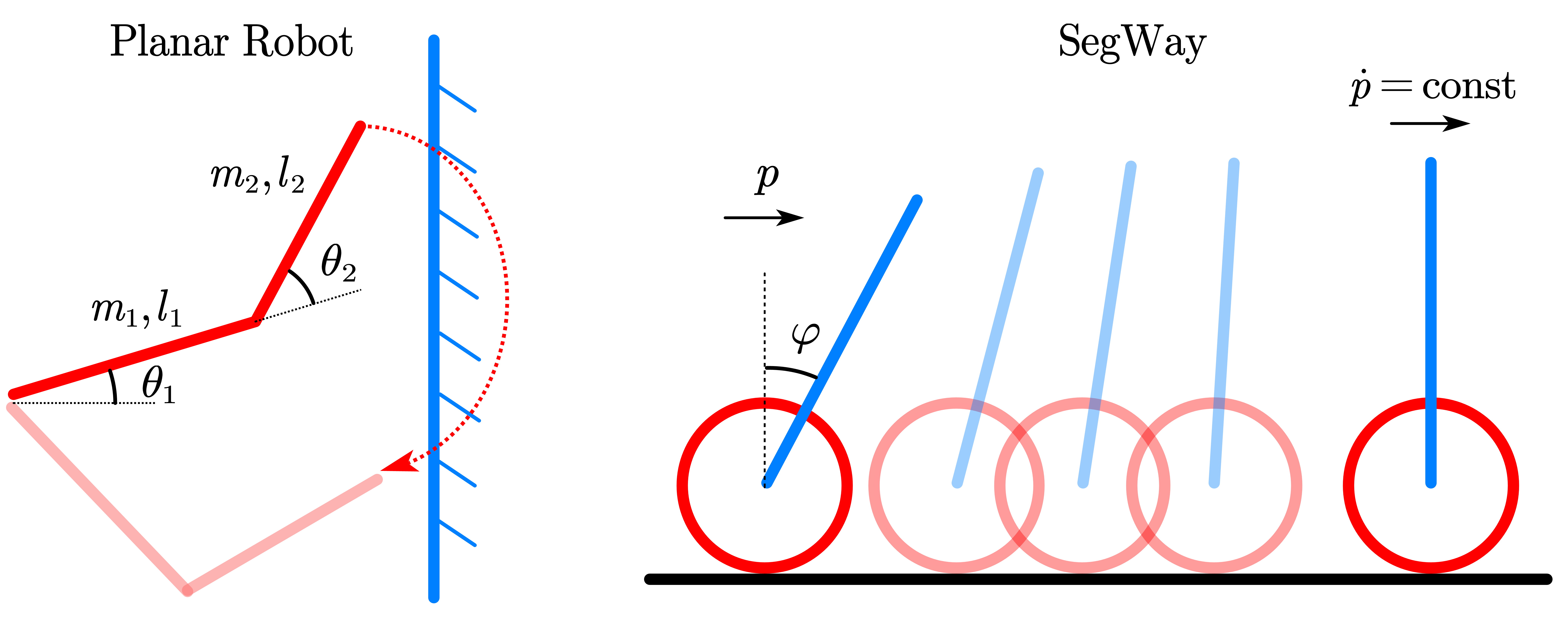}
    \caption{\small Illustration of the \re{planar robot} and Segway}
    \label{fig: robots}
\end{figure}


\subsection{\re{Planar Robot}}

We first test our method on a \re{planar robot} platform. We define the safety specification as not colliding with a vertical wall. Define the position and velocity of \re{planar robot}'s end-effector as $p_{\text{ee}} = [x_{\text{ee}}, y_{\text{ee}}]^T$ and $\dot{p}_{\text{ee}} = [\dot{x}_{\text{ee}}, \dot{y}_{\text{ee}}]^T$. Then the safety constraint $\phi_0 = -\Delta x = x_{\text{ee}} - x_{\text{wall}} < 0$ should always be satisfied. 
The detailed dynamic model can be found in \cref{apdx: SCARA_dynamics}.
We assume the payload $m_2$ (the mass of the second arm) is uncertain. Specifically, we assume $m_2\sim \mathcal{N} ( 1.0,0.3^2 ) \cdot I_{[ 0.1,1.9 ]}$. But after transformation, the dynamic model uncertainty is not Gaussian (the transformation involves multiplicative inverse and matrix inverse). 

\subsubsection{Feasibility}
\begin{figure}[tb]
\centering
\subfigure[$\phi_0$]{
\centering
\includegraphics[width=.33\linewidth]{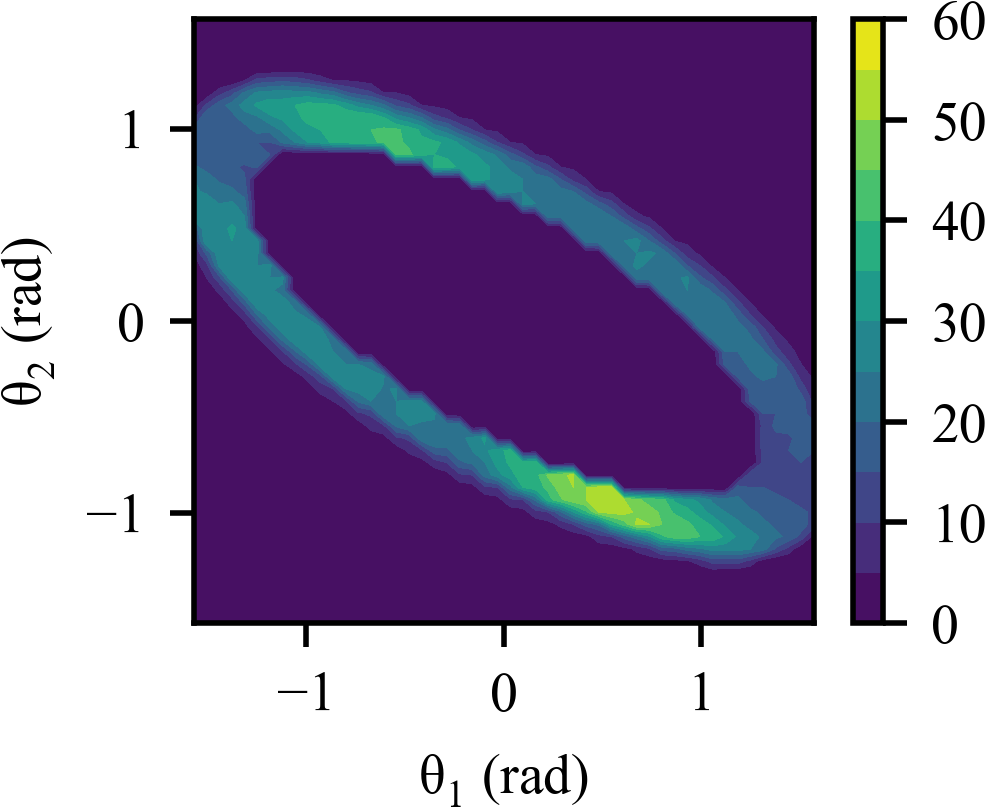}
}   \hspace{-3mm}
\subfigure[$\phi_h$]{
\centering
\includegraphics[width=.30\linewidth]{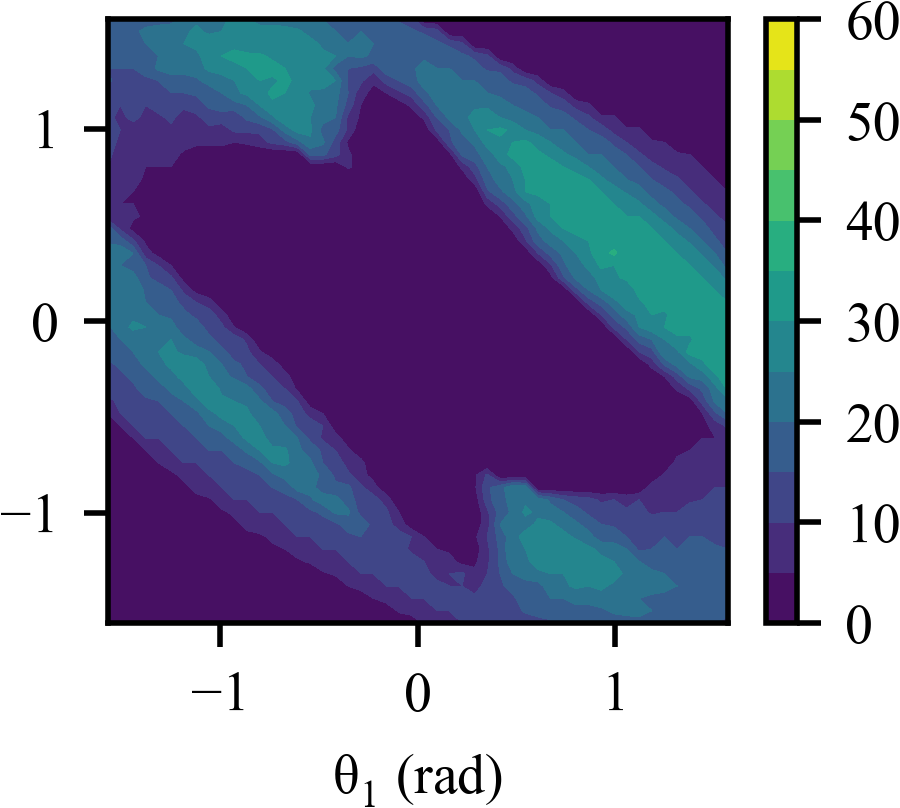}
}   \hspace{-3mm}
\subfigure[$\phi_l$]{
\centering
\includegraphics[width=.30\linewidth]{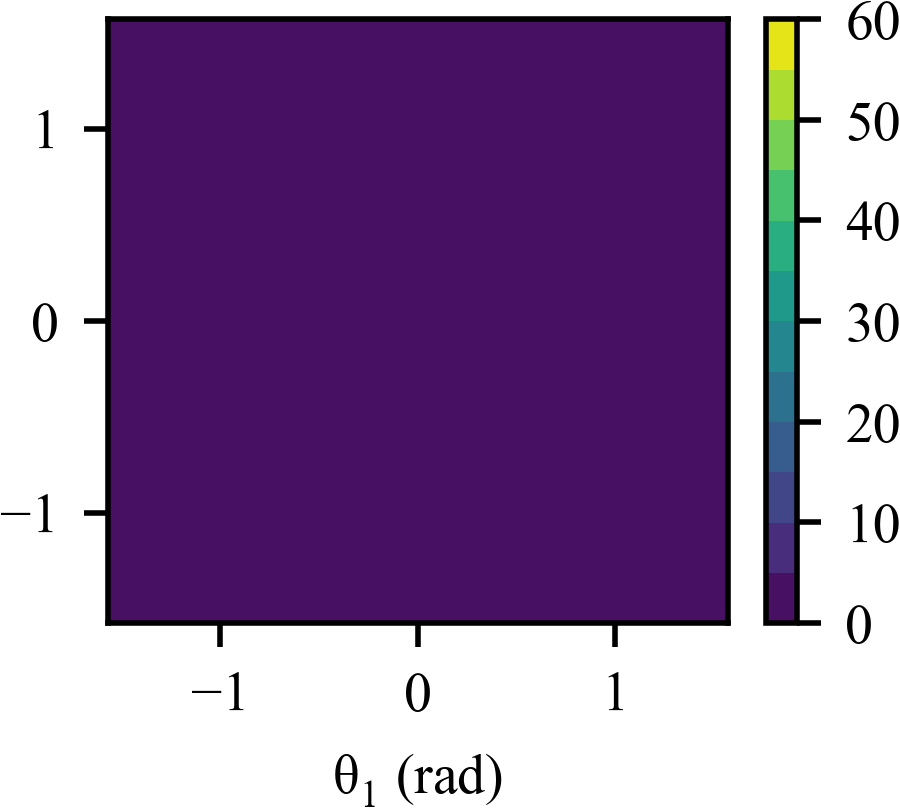}
}
\centering
\vspace{-10pt}
\caption{\small Distribution of infeasible states in the robot configuration space of three safety indexes. Each point of the graph represents a joint position. We sample 100 joint velocities at each point. The color denotes how many of them have no feasible robust safe control. (a) shows that the original safety index $\phi_0$ cannot ensure feasibility. (b) shows that a hand-designed safety index $\phi_h$ also has many infeasible states. (c) shows that our learned UR-SI $\phi_l$ ensures feasibility for all states even considering model uncertainty. \vspace{-10pt}}
\label{fig: UR-SI-learning}
\end{figure}

We first show that the safety index learning ensures feasibility.
To learn a UR-SI, we use the parameterized form proposed by~\cite{liu2014control}: $\phi = \max\{\phi_0,-x_{\text{wall}}^\alpha + x_{\text{ee}}^\alpha + k_v \dot{x}_{\text{ee}} + \beta$\}, where $\alpha, k_v, \beta$ are learnable parameters.
The search ranges for the parameters are: $\alpha \in ( 0.1,5.0 ) ,k_v\in ( 0.1,5.0 ) ,\beta \in ( 0.001,1.0 ) $. We set $x_{\text{wall}}=1.5$, and uniformly sample 250000 states in the whole state space. 
\Cref{fig: UR-SI-learning} compares $\phi_0$, a manually tuned safety index $\phi_h$ ($\alpha=1.0, k_v =  0.2, \beta = 0.0$), and a synthesized UR-SI $\phi_l$ with the Polytope RSSA solver.
($\alpha=0.57, k_v =  2.15, \beta = 0.072$). UR-SI achieves a $0$ infeasible rate.

\begin{figure}[tb]
    \vspace{5pt}
	\centering
	\subfigure[Case study 1. $\phi_0 < 0$, $\phi < 0$]{
	\begin{minipage}[c]{0.335\linewidth}
		\centering
		\hspace*{-0.35cm} 
		\vspace*{0.2cm}
		\includegraphics[width=1.0\linewidth]{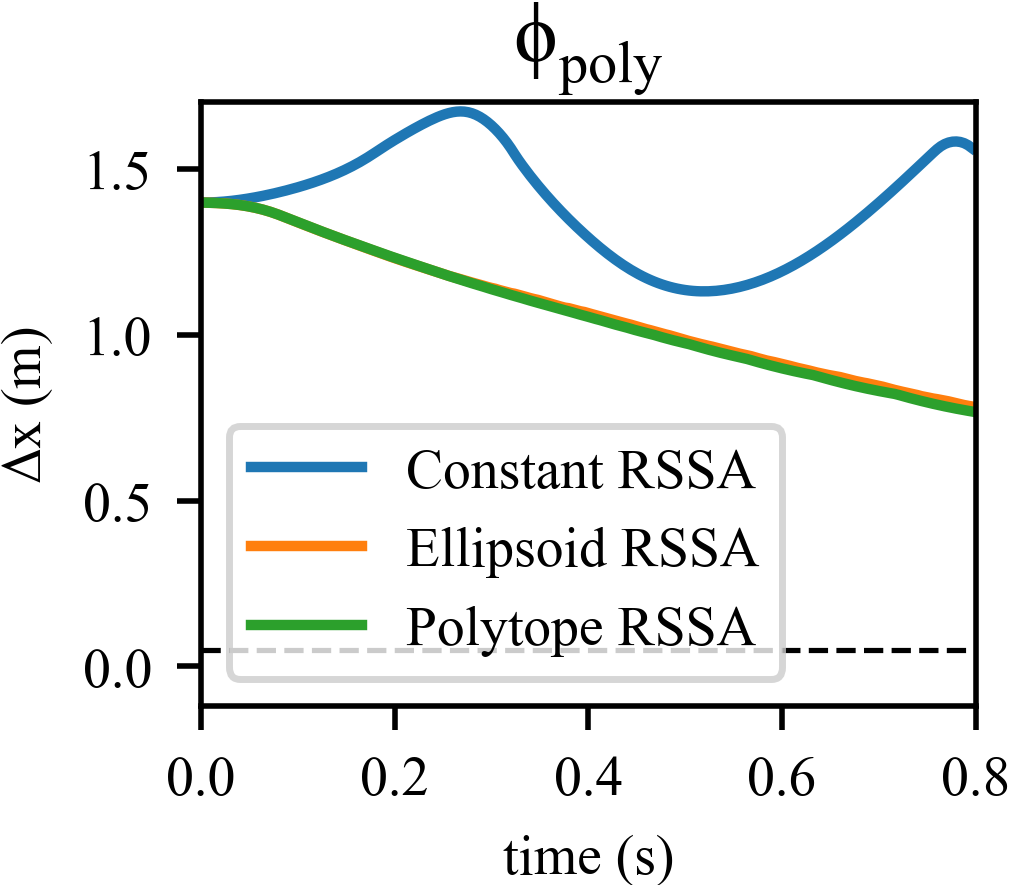}
		\label{fig: SCARA-RSSA-compare-phi_l_0__convex}
	\end{minipage} 
	\begin{minipage}[c]{0.31\linewidth}
		\centering
		\hspace*{-0.33cm}
		\vspace*{0.45cm}
		\includegraphics[width=1.0\linewidth]{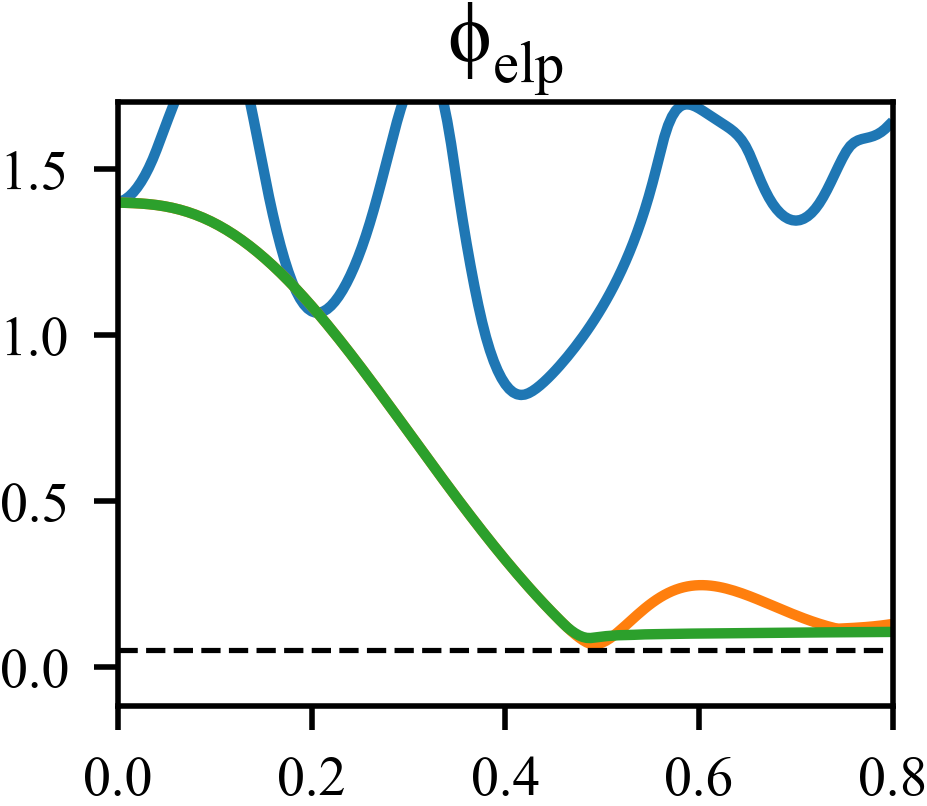}
		\label{fig: SCARA-RSSA-compare-phi_l_0__gaussian}
	\end{minipage} 
	\begin{minipage}[c]{0.31\linewidth}
		\centering
		\hspace*{-0.33cm}
		\vspace*{0.45cm}
		\includegraphics[width=1.0\linewidth]{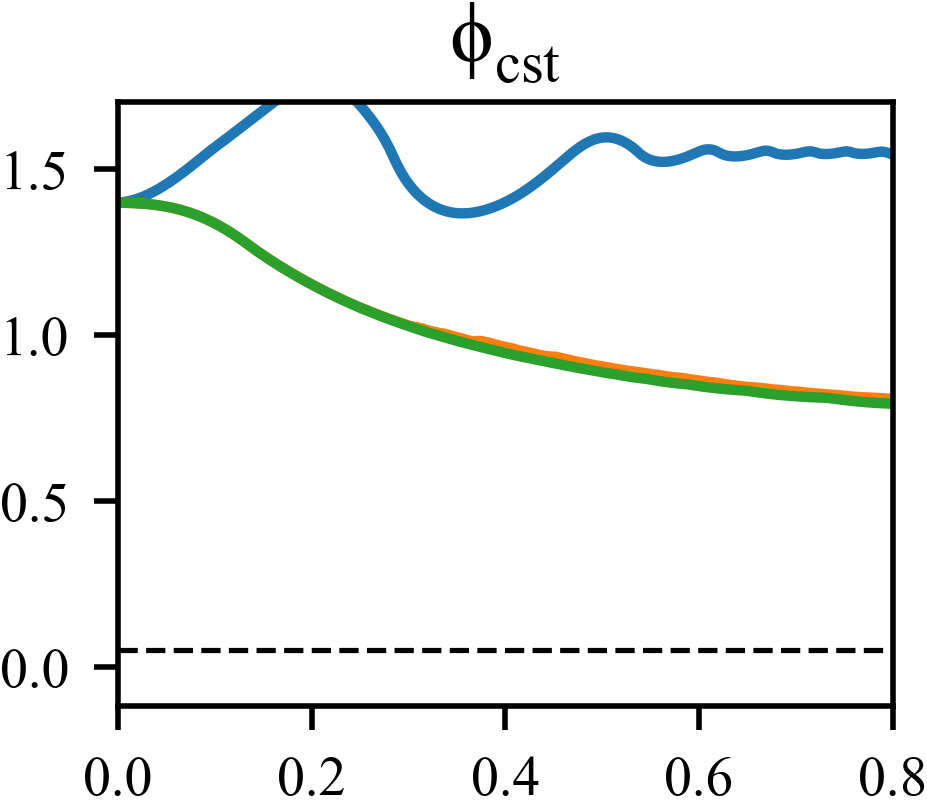}
		\label{fig: SCARA-RSSA-compare-phi_l_0__constant}
	\end{minipage} 
	}\label{fig: case_1}
	
	\subfigure[Case study 2. $\phi_0 < 0$, $\phi > 0$]{
	\begin{minipage}[c]{0.335\linewidth}
		\centering
            \hspace*{-0.35cm} 
		\vspace*{0.2cm}
		\includegraphics[width=1.0\linewidth]{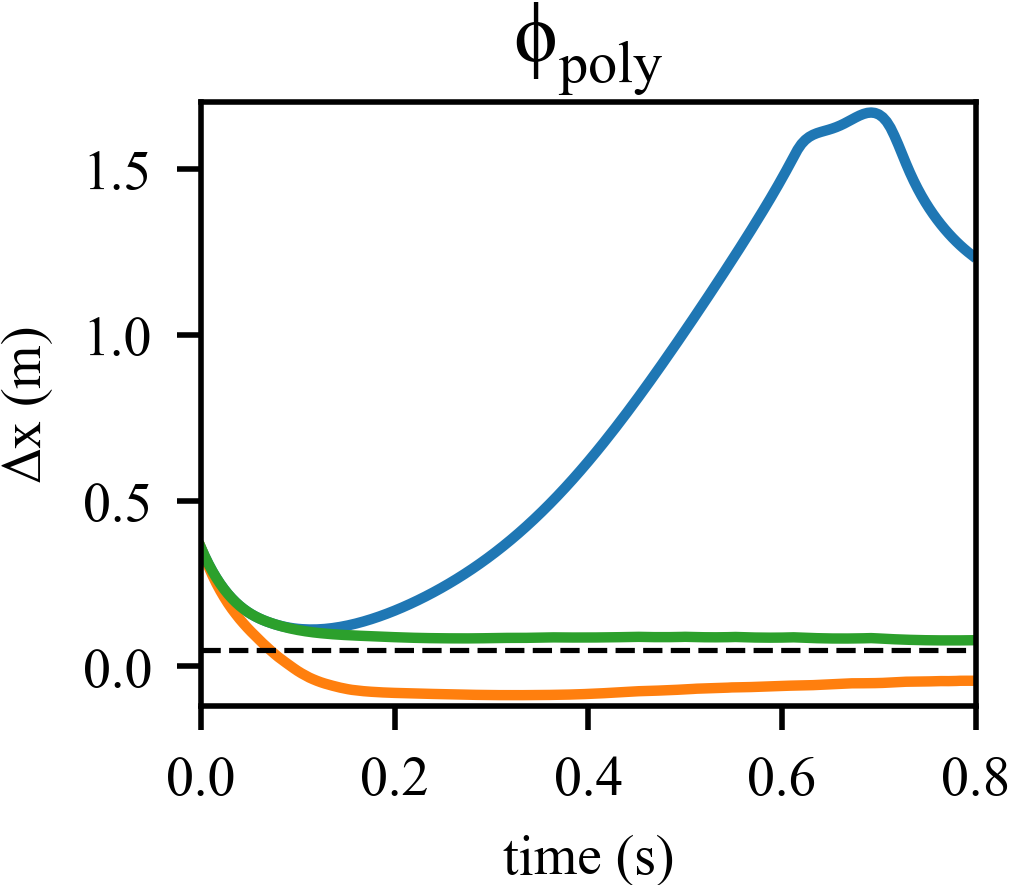}
		\label{fig: SCARA-RSSA-compare-phi_g_0__convex}
	\end{minipage} 
	\begin{minipage}[c]{0.31\linewidth}
		\centering
            \hspace*{-0.33cm}
		\vspace*{0.45cm}
		\includegraphics[width=1.0\linewidth]{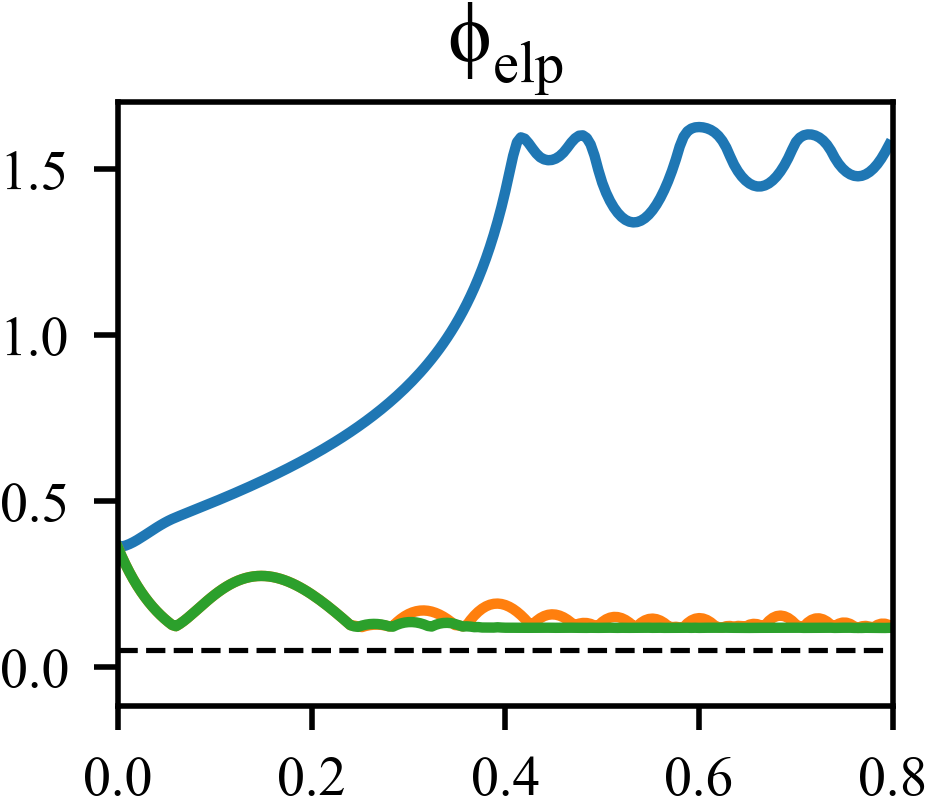}
		\label{fig: SCARA-RSSA-compare-phi_g_0__gaussian}
	\end{minipage} 
	\begin{minipage}[c]{0.31\linewidth}
		\centering
            \hspace*{-0.33cm}
		\vspace*{0.45cm}
		\includegraphics[width=1.0\linewidth]{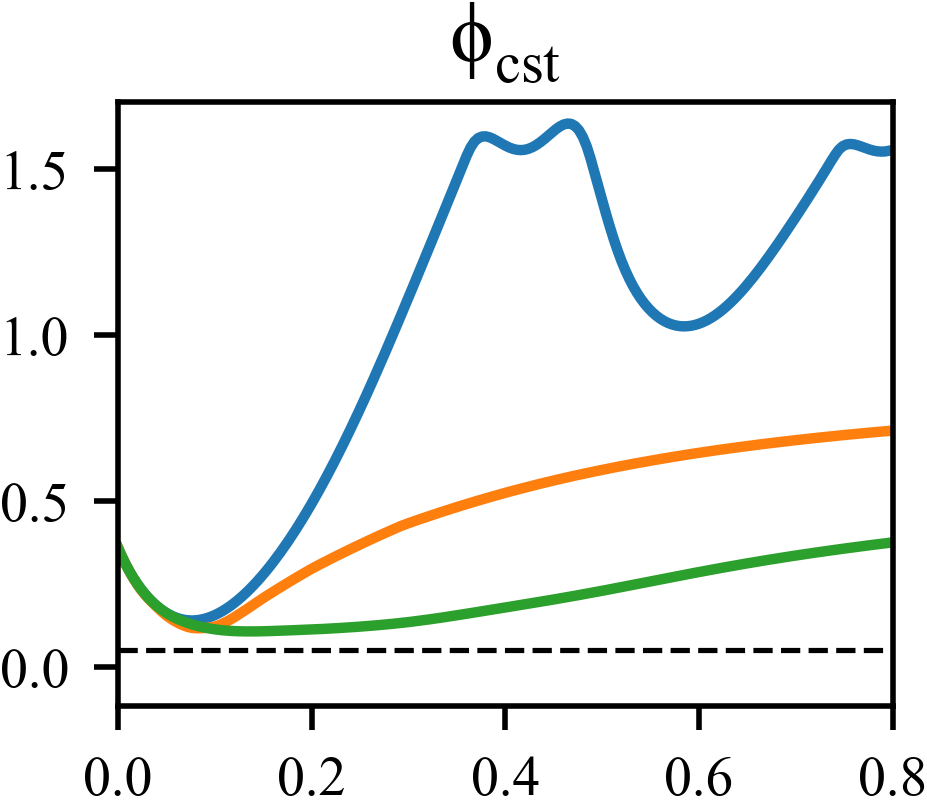}
		\label{fig: SCARA-RSSA-compare-phi_g_0__constant}
	\end{minipage} 
	}
    \vspace{-10pt}
    \caption{\small Comparison of RSSA solvers with three UR-SI and under two conditions. From left to right in each row, we learn a UR-SI based on different RSSA strategies: Polytope RSSA, Ellipsoid RSSA with $95\%$ confidence level, and Constant RSSA. In all cases, Polytope RSSA maintains safety and is less conservative than Constant RSSA. In the bottom-left graph, Ellipsoid RSSA fails to keep safe because the Gaussian assumption does not capture the uncertainty correctly. And if we use a large confidence level in Ellipsoid RSSA, e.g., $99\%$, CMA-ES cannot find a UR-SI. 
    The feasibility plot can be found in \cref{apdx: feasibility}.
    }
\label{fig: SCARA-CR-VS-GR}
\end{figure}

\begin{figure}[bt]

    \centering
    \subfigure[Uncertainty bounds]{
    \begin{minipage}[c]{0.46\linewidth}
    \centering\includegraphics[width=1\linewidth]{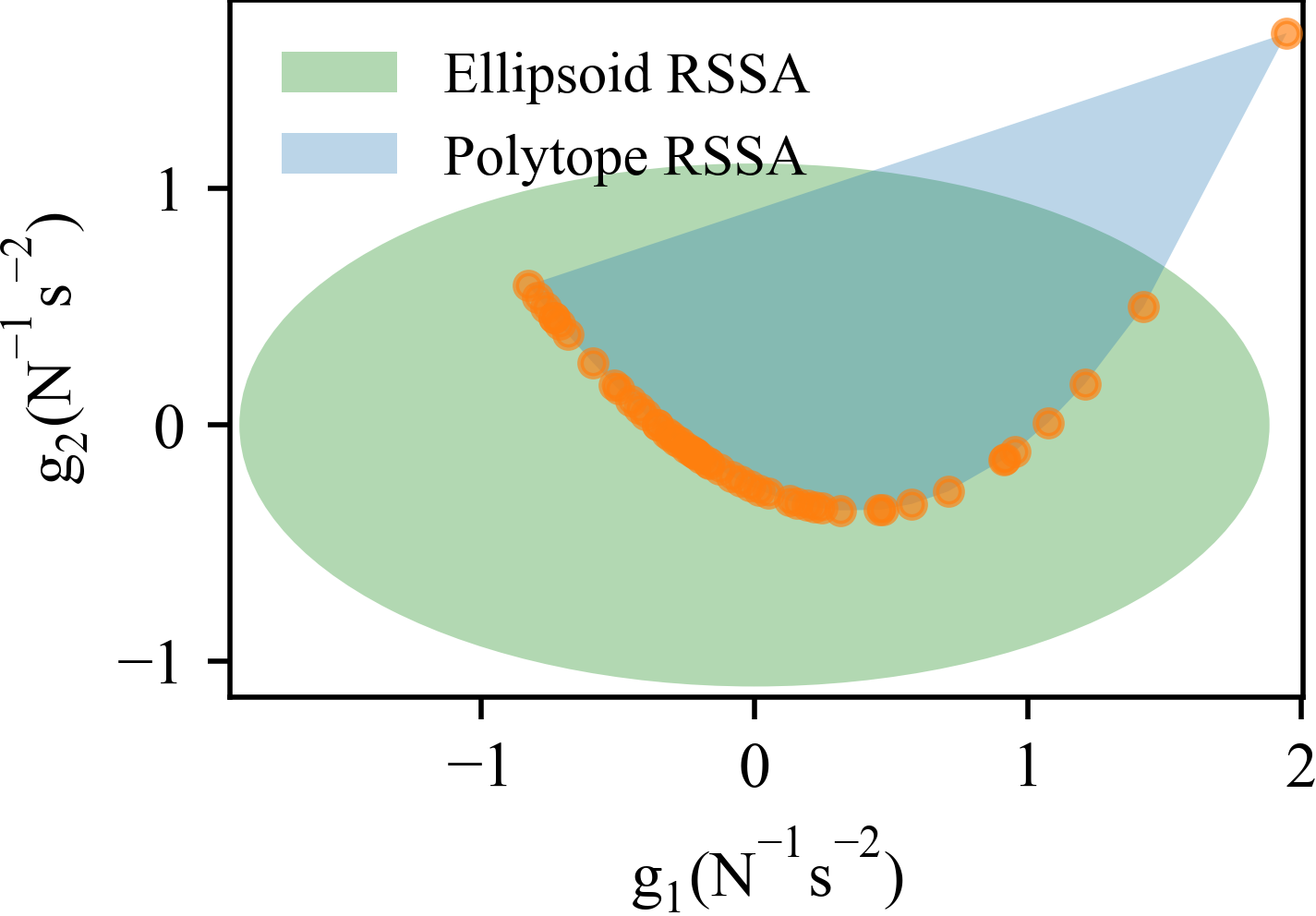}
    \vspace{-2mm}
    \end{minipage}
    }
    \subfigure[Safe control set]{
    \begin{minipage}[c]{0.46\linewidth}
    \centering
    \includegraphics[width=1\linewidth]{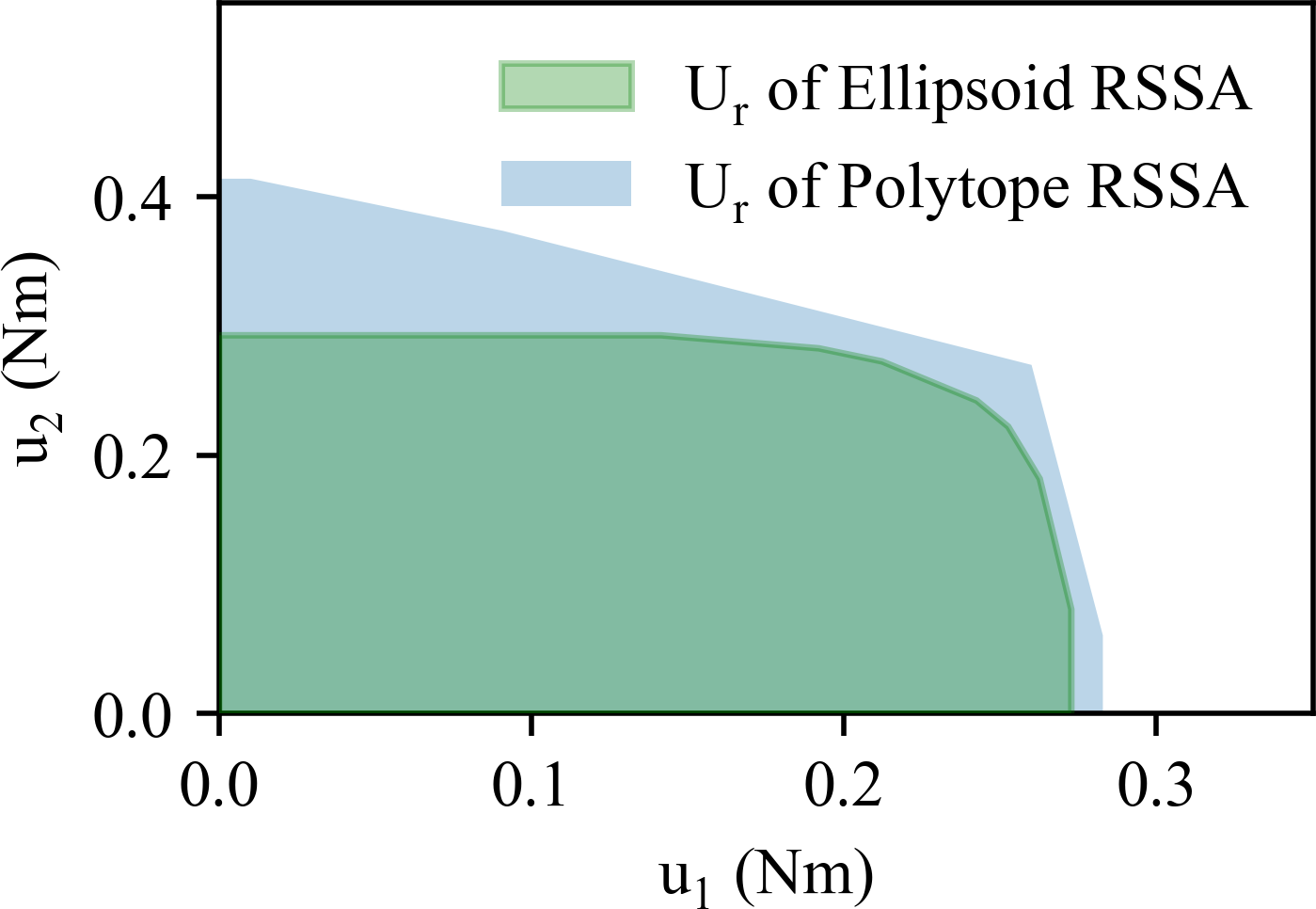}
    \vspace{-2mm}
    \end{minipage}
    }   
    \vspace{-10pt}
    \caption{\small (a) shows polytope \re{(CSIP)} and ellipsoid \re{(SOCP)} uncertainty bounds for a 2D slice of $g(x)$. Orange points represent dynamic model samples. The ellipsoid corresponds to the $3\sigma$ confidence region ($99.7\%$) of the fitted Gaussian distribution. It still does not cover the true dynamic model distribution, while the polytope bound does. (b) shows the corresponding robust safe control sets $U_r$. Polytope RSSA has a larger $U_r$ because it is less conservative.\vspace{-10pt}}
    \label{fig: Uncertainty_bounds_and_safe_control_set}
\end{figure}

\subsubsection{Robust safe control solver}

We compare the performance of our solvers Polytope RSSA and Ellipsoid RSSA, with a popular method: constant bounded robust safe control (Constant RSSA)~\cite{cosner2021MeasurementRobust,nguyen2022Robust,brunke2022Barrier}, which considers the uncertainty as residual dynamics, and uses a constant to bound the residual time derivative of $\phi$: $\dot \phi_{\text{res}}$ caused by uncertainties.







We construct the uncertainty bounds for both Ellipsoid RSSA and Polytope RSSA by sampling methods. We first get 50 dynamic model samples by sampling $m_2$ from the distribution and execute the nonlinear dynamics transformation. Then we either fit a Gaussian distribution for Ellipsoid RSSA or construct a polytope for Polytope RSSA.

For a fair comparison, we learned a UR-SI for each solver, denoted by $\phi_{\text{poly}}$, $\phi_{\text{elp}}$ and $\phi_{\text{cst}}$.
With a true yet unknown $m_2 = 0.5$, we plot trajectories of $\Delta x$ of all the solvers with the three UR-SIs under two different initial conditions: 
1. when $\phi_0 < 0$ and $\phi < 0$, \cref{fig: SCARA-CR-VS-GR} (a) shows that all the methods ensures safety, but as for the conservativeness, Polytope RSSA $<$ Ellipsoid RSSA $<$ Constant RSSA.  2. when $\phi_0 < 0$ but $\phi > 0$, \re{\cref{fig: SCARA-CR-VS-GR} (b) shows that Ellipsoid RSSA violates the safety constraint; To explain it in detail, we visualize the uncertainty bounds and corresponding safe control sets in \cref{fig: Uncertainty_bounds_and_safe_control_set}. It shows that even the $99.7\%$ Gaussian confidence bound does not cover the uncertainty correctly, leading to potential safety violations. It demonstrates that existing methods assuming Gaussian uncertainties may be unreliable when the uncertainty is not Gaussian. Besides, we can see that the SOCP formulation is not as tight as the CSIP formulation in general. } Note that Ellipsoid RSSA also can guarantee safety if the bound correctly covers the uncertainty, and the bound does not have to be constructed from Gaussian confidence intervals.


\subsubsection{Unmodeled uncertainties}
We study the forward invariance set change under unmodeled uncertainties. We test several $m_2$ that are out of the modeled uncertainty bound $[0.1,1.9]$ in safety index learning and record $\phi_{\max}$: the largest $\phi$ of many sampled trajectories, as an indicator of the forward invariant set boundary. \Cref{tab: FI_set} shows that the forward invariant set grows incrementally with the uncertainty level, and $\phi_{\max}$ is always below the theoretical bound predicted in \cref{thm: FI_set}.

\begin{table}[htbp]
    \centering
    \caption{\small Forward invariance with unmodeled uncertainties.\vspace{-15pt}}
    \footnotesize
    \begin{tabular}{cccccc}
    \toprule
    $m_2$  & 0.005 & 0.08  & 1.0     & 2.0     & 3.0\\
    \midrule
    $\phi_{\max}$ & 32.09 & 0.41  & -0.0013 & 0.92  & 1.51\\
    \midrule
    Theoretical bound & 256.4 & 3.51  & 0.0     & 1.07  & 8.75\\
    \bottomrule
    \end{tabular}%
  \label{tab: FI_set}%
\end{table}%

\subsection{Segway}

To highlight the optimality and efficiency of Ellipsoid RSSA in case of Gaussian uncertainty. We also test our method on a realistic Segway model.
We consider a tracking task with a safety specification on the tilt angle: $\phi_0 = |\varphi| - 0.1$. We consider a safety index $\phi = \max\{\phi_0,-0.1^\alpha + |\varphi|^\alpha + k_v\text{sign}(\varphi)  \dot{\varphi} + \beta$\}, where $\alpha, k_v, \beta$ are learnable parameters. The nominal controller is designed to maintain $\dot{p}$ at $1m/s$. 
The detailed dynamics is shown in \cref{apdx: Segway_dynamics}.
We assume the motor torque constant $K_m$ is a truncated Gaussian distribution: $K_m\sim \mathcal{N} ( 2.524,0.3^2 ) \cdot I_{[ 1.624,3.424 ]}
$. After transformation, the dynamic model uncertainty is a state-dependent Gaussian distribution.

\subsubsection{Robust safe control solvers}

\begin{figure}[tbh]
    \vspace{5pt}
    \centering
    \subfigure[]{
    \centering
    \includegraphics[width=0.45\linewidth]{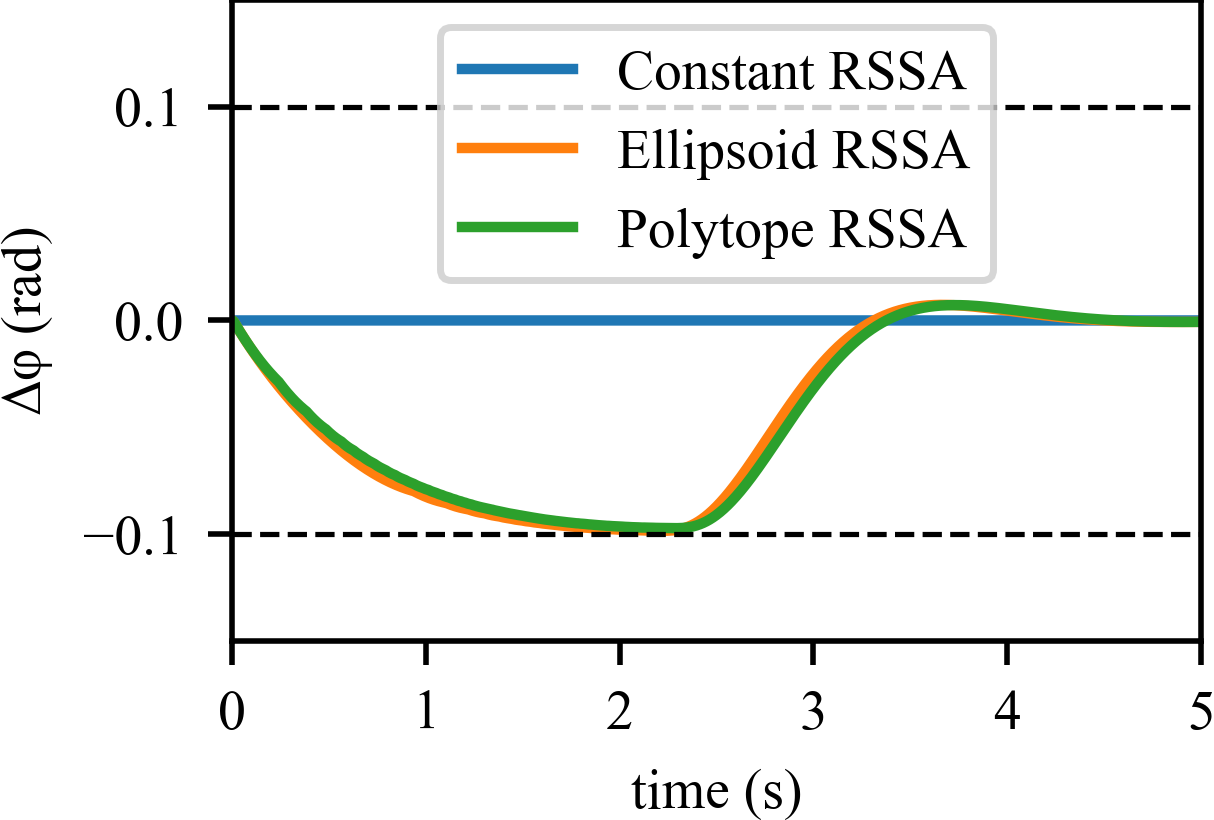}
    }
    \subfigure[]{
    \centering
    \includegraphics[width=0.43\linewidth]{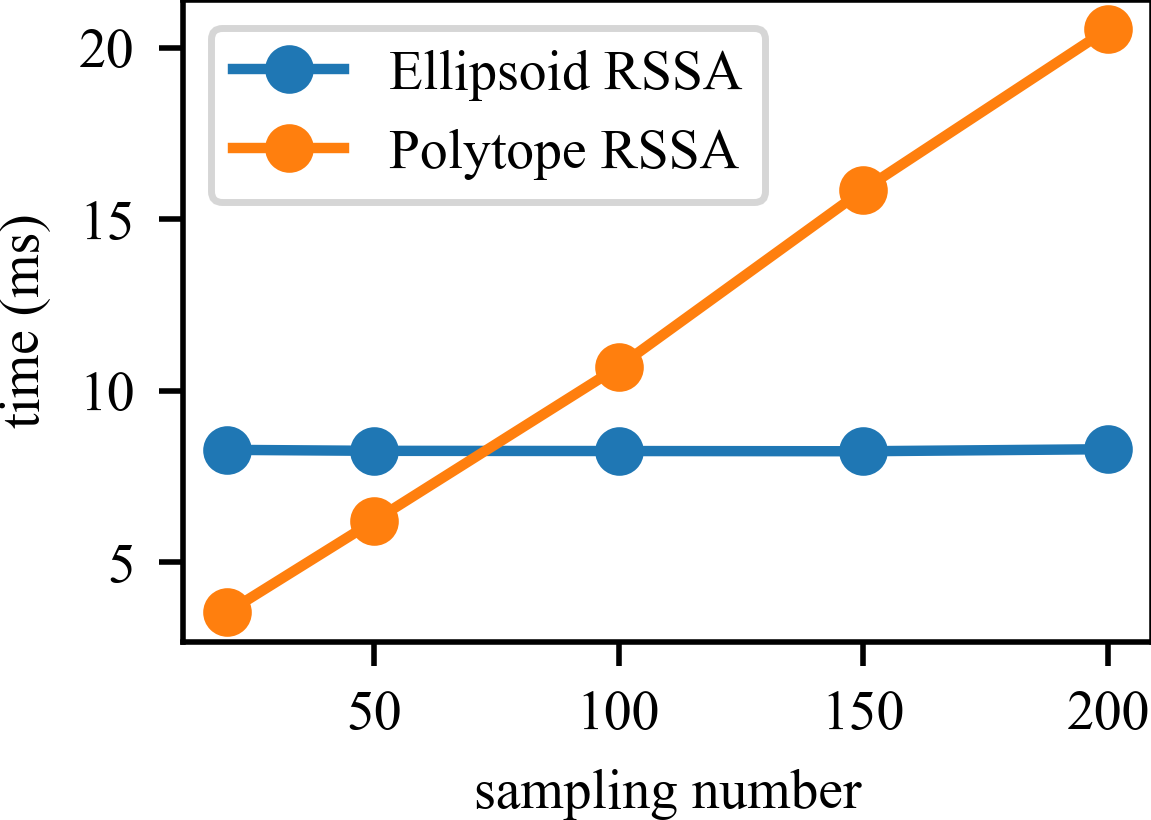}
    }
    \vspace{-10pt}
    \caption{\small (a) Comparing three different RSSA strategies in Segway. Both Polytope RSSA and Ellipsoid RSSA maintain safety. However, Constant RSSA always forces $u$ to zero (not moving) because of conservativeness. (b) Comparison of the average computation time of Polytope RSSA and Ellipsoid RSSA.\vspace{-15pt}}
    \label{fig: SegWay-analysis}
\end{figure}

We can efficiently compute the correct ellipsoid bounds for Ellipsoid RSSA because the dynamic model follows Gaussian distributions. But for Polytope RSSA, we still rely on the sampling method to construct the bound. As shown in \cref{fig: SegWay-analysis} (a), they both ensure safety and are non-conservative. But the computation time of Ellipsoid RSSA does not grow with the number of samples as shown in \cref{fig: SegWay-analysis} (b), which is a significant advantage when the dimensions are high and require more samples. The constant RSSA is overly conservative because $\dot \phi_{\text{res}}$ is too large.

\section{Discussion}

In this work, we proposed a general framework of robust safe control for uncertain dynamic models. The future work includes studying complete methods for safety index synthesis, generalizing the method to the case that $f$ and $g$ have correlations, and generalizing the method to high-dimensional applications that require a large number of samples to learn the safety index and construct bounds.




\bibliographystyle{IEEEtran}
\bibliography{reference}

\begin{thebibliography}{10}
\providecommand{\url}[1]{#1}
\csname url@samestyle\endcsname
\providecommand{\newblock}{\relax}
\providecommand{\bibinfo}[2]{#2}
\providecommand{\BIBentrySTDinterwordspacing}{\spaceskip=0pt\relax}
\providecommand{\BIBentryALTinterwordstretchfactor}{4}
\providecommand{\BIBentryALTinterwordspacing}{\spaceskip=\fontdimen2\font plus
\BIBentryALTinterwordstretchfactor\fontdimen3\font minus
  \fontdimen4\font\relax}
\providecommand{\BIBforeignlanguage}[2]{{%
\expandafter\ifx\csname l@#1\endcsname\relax
\typeout{** WARNING: IEEEtran.bst: No hyphenation pattern has been}%
\typeout{** loaded for the language `#1'. Using the pattern for}%
\typeout{** the default language instead.}%
\else
\language=\csname l@#1\endcsname
\fi
#2}}
\providecommand{\BIBdecl}{\relax}
\BIBdecl

\bibitem{liu2014control}
C.~Liu and M.~Tomizuka, ``Control in a safe set: Addressing safety in
  human-robot interactions,'' in \emph{ASME 2014 Dynamic Systems and Control
  Conference}.\hskip 1em plus 0.5em minus 0.4em\relax American Society of
  Mechanical Engineers Digital Collection, 2014.

\bibitem{wei2019safe}
T.~Wei and C.~Liu, ``Safe control algorithms using energy functions: A uni ed
  framework, benchmark, and new directions,'' in \emph{2019 IEEE 58th
  Conference on Decision and Control (CDC)}.\hskip 1em plus 0.5em minus
  0.4em\relax IEEE, 2019, pp. 238--243.

\bibitem{wajid2022Formal}
R.~Wajid, A.~U. Awan, and M.~Zamani, ``Formal {{Synthesis}} of {{Safety
  Controllers}} for {{Unknown Stochastic Control Systems}} using {{Gaussian
  Process Learning}},'' in \emph{Proceedings of {{The}} 4th {{Annual Learning}}
  for {{Dynamics}} and {{Control Conference}}}.\hskip 1em plus 0.5em minus
  0.4em\relax {PMLR}, May 2022, pp. 624--636.

\bibitem{taylor2021towards}
A.~J. Taylor, V.~D. Dorobantu, S.~Dean, B.~Recht, Y.~Yue, and A.~D. Ames,
  ``Towards robust data-driven control synthesis for nonlinear systems with
  actuation uncertainty,'' in \emph{2021 60th IEEE Conference on Decision and
  Control (CDC)}.\hskip 1em plus 0.5em minus 0.4em\relax IEEE, 2021, pp.
  6469--6476.

\bibitem{castaneda2021pointwise}
F.~Casta{\~n}eda, J.~J. Choi, B.~Zhang, C.~J. Tomlin, and K.~Sreenath,
  ``Pointwise feasibility of gaussian process-based safety-critical control
  under model uncertainty,'' in \emph{2021 60th IEEE Conference on Decision and
  Control (CDC)}.\hskip 1em plus 0.5em minus 0.4em\relax IEEE, 2021, pp.
  6762--6769.

\bibitem{buch2021robust}
J.~Buch, S.-C. Liao, and P.~Seiler, ``Robust control barrier functions with
  sector-bounded uncertainties,'' \emph{IEEE Control Systems Letters}, vol.~6,
  pp. 1994--1999, 2021.

\bibitem{garg2021robust}
K.~Garg and D.~Panagou, ``Robust control barrier and control lyapunov functions
  with fixed-time convergence guarantees,'' in \emph{2021 American Control
  Conference (ACC)}.\hskip 1em plus 0.5em minus 0.4em\relax IEEE, 2021, pp.
  2292--2297.

\bibitem{jankovic2018robust}
M.~Jankovic, ``Robust control barrier functions for constrained stabilization
  of nonlinear systems,'' \emph{Automatica}, vol.~96, pp. 359--367, Oct. 2018.

\bibitem{cosner2021MeasurementRobust}
R.~K. Cosner, A.~W. Singletary, A.~J. Taylor, T.~G. Molnar, K.~L. Bouman, and
  A.~D. Ames, ``Measurement-{{Robust Control Barrier Functions}}: {{Certainty}}
  in {{Safety}} with {{Uncertainty}} in {{State}},'' in \emph{2021
  {{IEEE}}/{{RSJ International Conference}} on {{Intelligent Robots}} and
  {{Systems}} ({{IROS}})}, Sep. 2021, pp. 6286--6291.

\bibitem{nguyen2022Robust}
Q.~Nguyen and K.~Sreenath, ``Robust {{Safety-Critical Control}} for {{Dynamic
  Robotics}},'' \emph{IEEE Transactions on Automatic Control}, vol.~67, no.~3,
  pp. 1073--1088, Mar. 2022.

\bibitem{brunke2022Barrier}
L.~Brunke, S.~Zhou, and A.~P. Schoellig, ``Barrier {{Bayesian Linear
  Regression}}: {{Online Learning}} of {{Control Barrier Conditions}} for
  {{Safety-Critical Control}} of {{Uncertain Systems}},'' in \emph{Proceedings
  of {{The}} 4th {{Annual Learning}} for {{Dynamics}} and {{Control
  Conference}}}.\hskip 1em plus 0.5em minus 0.4em\relax {PMLR}, May 2022, pp.
  881--892.

\bibitem{grover2022control}
J.~S. Grover, C.~Liu, and K.~Sycara, ``Control barrier functions-based
  semi-definite programs (cbf-sdps): Robust safe control for dynamic systems
  with relative degree two safety indices,'' \emph{arXiv preprint
  arXiv:2208.12252}, 2022.

\bibitem{liu2016algorithmic}
C.~Liu and M.~Tomizuka, ``Algorithmic safety measures for intelligent
  industrial co-robots,'' in \emph{2016 IEEE International Conference on
  Robotics and Automation (ICRA)}.\hskip 1em plus 0.5em minus 0.4em\relax IEEE,
  2016, pp. 3095--3102.

\bibitem{noren2021safe}
C.~Noren, W.~Zhao, and C.~Liu, ``Safe {{Adaptation}} with {{Multiplicative
  Uncertainties Using Robust Safe Set Algorithm}},'' \emph{IFAC-PapersOnLine},
  vol.~54, no.~20, pp. 360--365, Jan. 2021.

\bibitem{wei2022safe}
T.~Wei and C.~Liu, ``Safe control with neural network dynamic models,'' in
  \emph{Learning for Dynamics and Control Conference}.\hskip 1em plus 0.5em
  minus 0.4em\relax PMLR, 2022, pp. 739--750.

\bibitem{liu2015safe}
C.~Liu and M.~Tomizuka, ``Safe exploration: Addressing various uncertainty
  levels in human robot interactions,'' in \emph{2015 American Control
  Conference (ACC)}.\hskip 1em plus 0.5em minus 0.4em\relax IEEE, 2015, pp.
  465--470.

\bibitem{gustafson1973numerical}
S.~Gustafson and K.~Kortanek, ``Numerical treatment of a class of semi-infinite
  programming problems,'' \emph{Naval Research Logistics Quarterly}, vol.~20,
  no.~3, pp. 477--504, 1973.

\end{thebibliography}

\addtolength{\textheight}{-12cm}   


\appendix

\subsection{SCARA}\label{apdx: SCARA_dynamics}

Define joint positions of the robot arm as $\theta = [\theta_1, \theta_2]^T$ and joint velocities as $\dot{\theta} = [\dot{\theta}_1, \dot{\theta}_2]^T$. 
Here we assume all states $x=[\theta_1, \theta_2, \dot{\theta}_1, \dot{\theta}_2]^T$ and control inputs $u$ are bounded: $X: [ -\frac{\pi}{2},\frac{\pi}{2} ] \times [ -\frac{\pi}{2},\frac{\pi}{2} ] \times [ -2,2 ] \times [ -2,2 ] $ and $U : [-20,20]\times [-20,20]$.

The control-affine dynamic model for SCARA is as follows: 
\begin{align}
    \frac{d}{dt}\left[ \begin{array}{c}	\theta\\	\dot{\theta}\\\end{array} \right] =\left[ \begin{array}{c}	\dot{\theta}\\	-M\left( \theta \right) ^{-1}H\left( \theta ,\dot{\theta} \right)\\\end{array} \right] +\left[ \begin{array}{c}	0\\	M\left( \theta \right) ^{-1}\\\end{array} \right] u \label{eq: scara_dynamic_model}
\end{align}

where $M(\theta)$ is the mass matrix and $H(\theta, \dot{\theta})$ is the Coriolis matrix: 
\begin{align}
    M\left( \theta _1,\theta _2 \right) =\left[ \begin{matrix}	2A+2B+2C\cos \left( \theta _2 \right)&		2B+C\cos \left( \theta _2 \right)\\	2B+C\cos \left( \theta _2 \right)&		2B\\\end{matrix} \right]
\end{align}
\begin{align}
    H\left( \theta _1,\theta _2,\dot{\theta}_1,\dot{\theta}_2 \right) =\left[ \begin{array}{c}	-C\sin \theta _2\cdot \left( 2\dot{\theta}_1+\dot{\theta}_2 \right) \dot{\theta}_2\\	C\sin \theta _2\cdot \dot{\theta}_{1}^{2}\\\end{array} \right] 
\end{align}

And $A, B, C$ only depend on the robot arm's masses and lengths: 
\begin{align}
    \begin{cases}	A=\frac{1}{6}m_1l_{1}^{2}+\frac{1}{2}m_2l_{1}^{2}\\	B=\frac{1}{6}m_2l_{2}^{2}\\	C=\frac{1}{2}m_2l_1l_2\\\end{cases}
\end{align}

\subsection{Segway}\label{apdx: Segway_dynamics}


Given wheel's position $p$ and frame's tilt angle $\varphi$, we define $q = [p, \varphi]^T$ and $\dot{q} = [\dot{p}, \dot{\varphi}]^T$. SegWay's dynamic model can be written as
\begin{align}
    \frac{d}{dt}\left[ \begin{array}{c}
        	q\\
        	\dot{q}\\
        \end{array} \right] =\left[ \begin{array}{c}
        	\dot{q}\\
        	-M\left( q \right) ^{-1}H\left( q,\dot{q} \right)\\
        \end{array} \right] +\left[ \begin{array}{c}
        	0\\
        	M\left( q \right) ^{-1}B\\
        \end{array} \right] u
\end{align}

where $M(q)$, $H(q, \dot{q})$, $B$ and $b_t$ are defined as follows: 
\begin{align}
    M\left( q \right) =\left[ \begin{matrix}
        	m_0&		mL\cos \left( \varphi \right)\\
        	mL\cos \left( \varphi \right)&		J_0\\
        \end{matrix} \right] 
\end{align}
\begin{align}
    H\left( q,\dot{q} \right) =\left[ \begin{array}{c}
        	-mL\sin \left( \varphi \right) \dot{\varphi}^2+\frac{b_t}{R}\left( \dot{p}-R\dot{\varphi} \right)\\
        	-mgL\sin \left( \varphi \right) -b_t\left( \dot{p}-R\dot{\varphi} \right)\\
        \end{array} \right] 
\end{align}
\begin{align}
    B=\left[ \begin{array}{c}
        	\frac{K_m}{R}\\
        	-K_m\\
        \end{array} \right]
\end{align}
\begin{align}
    b_t=K_m\frac{K_b}{R}
\end{align}

\begin{figure}[tbh]
	\centering
	\subfigure[Case study 1. $\phi_0 < 0$, $\phi < 0$]{
	    \includegraphics[width=\textwidth]{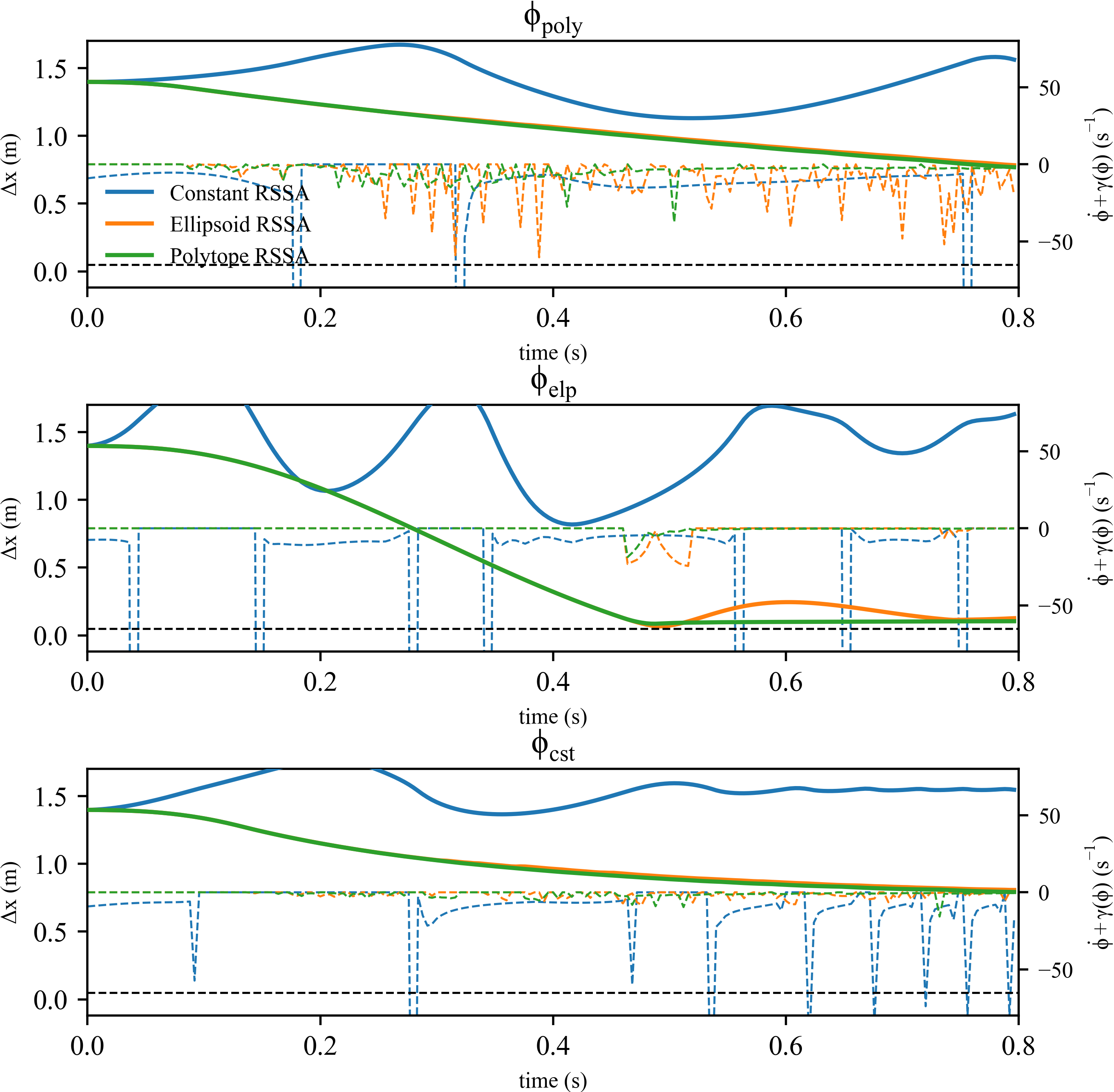}
	}
	\subfigure[Case study 2. $\phi_0 < 0$, $\phi > 0$]{
	      \includegraphics[width=\textwidth]{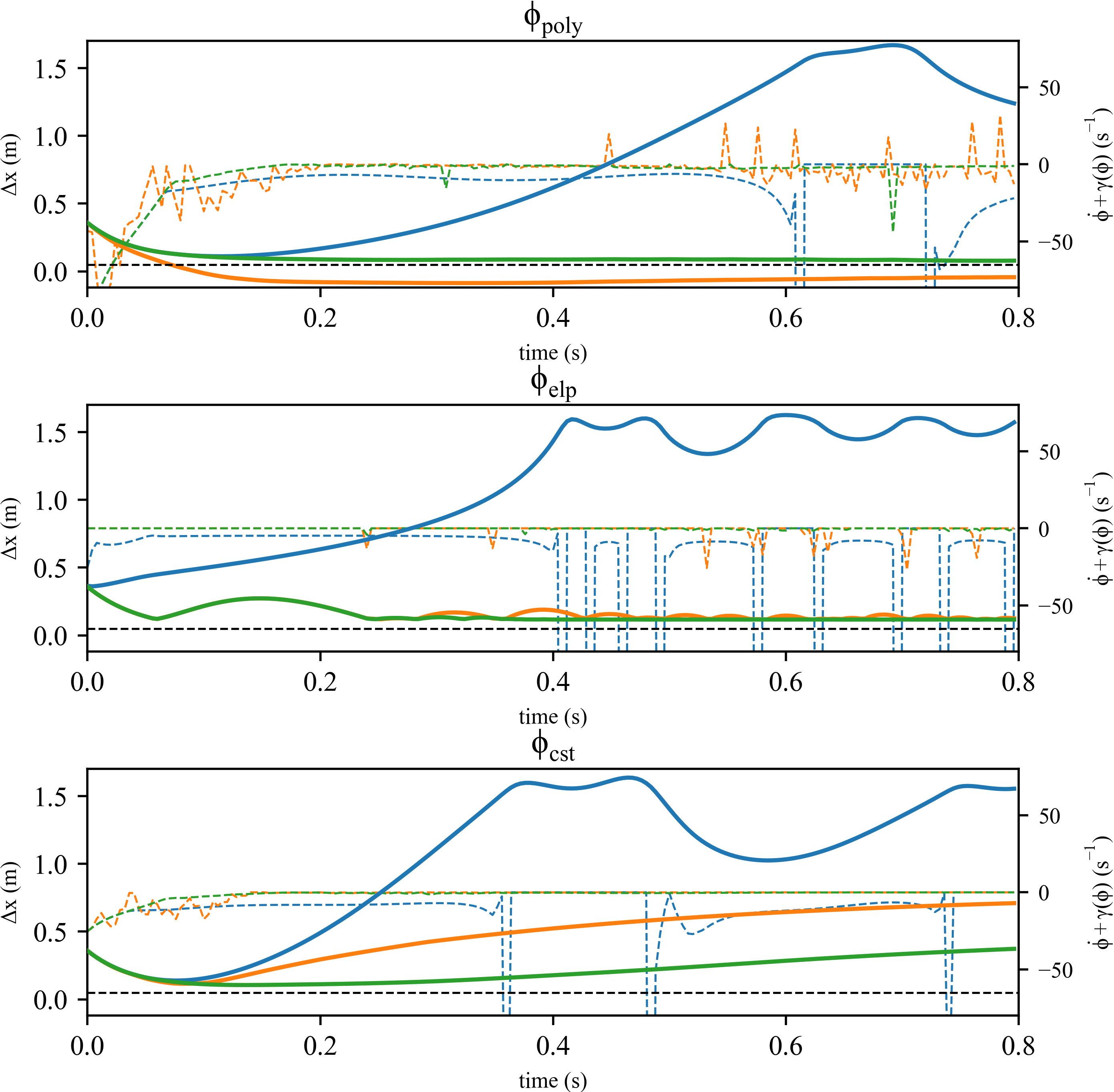} 
	}
    \caption{\small $\Delta x\sim \dot{\phi}+\gamma \left( \phi \right) $}
\label{fig: SCARA-CR-VS-GR-add-dot-phi}
\end{figure}

\pagebreak
\rre{
\subsection{Feasibility Plot}\label{apdx: feasibility}
\Cref{fig: SCARA-CR-VS-GR-add-dot-phi} shows the trajectory of $\phi_0$ and $\dot\phi + \gamma(\phi)$. Solid lines represent the value of $-\phi_0$, and the dashed lines represent $\dot\phi + \gamma(\phi)$. We can see from the figure that the Polytope RSSA safety index is persistently feasible ($\dot \phi + \gamma(\phi)$ always $<0$) while the Ellipsoid RSSA one is not.
}

\end{document}